\RequirePackage{fix-cm}
\documentclass[smallcondensed]{svjour3}     % onecolumn (ditto)
\smartqed  % flush right qed marks, e.g. at end of proof
\usepackage{graphicx}
\usepackage{amssymb,amsfonts}
\usepackage{url,epsfig,amsmath,paralist}
\usepackage[ruled,vlined]{algorithm2e}
\SetKwInOut{KwInit}{Initialization}
\DeclareMathOperator*{\argmax}{arg\,max}

\usepackage{xcolor}
\usepackage{booktabs}
\usepackage{graphicx}
\usepackage{multirow}
\usepackage{epstopdf}
\usepackage{float}
\usepackage{subfig}
\definecolor{g}{RGB}{0, 150, 0}

\begin{document}

\title{Mining Drifting Data Streams on a Budget: Combining Active Learning with Self-Labeling
}

\titlerunning{Mining Drifting Data Streams on a Budget}       

\author{\L{}ukasz Korycki         \and
        Bartosz Krawczyk %etc.
}

\institute{\L{}ukasz Korycki \at
              Virginia Commonwealth University, 401 West Main Street, Richmond, VA 23284-3019, USA\\
              \email{koryckil@vcu.edu}           %  \\
%             \emph{Present address:} of F. Author  %  if needed
           \and
           Bartosz Krawczyk \at
              Virginia Commonwealth University, 401 West Main Street, Richmond, VA 23284-3019, USA\\
              \email{bkrawczyk@vcu.edu}
}

\date{}
% The correct dates will be entered by the editor

\maketitle

\begin{abstract}
Mining data streams poses a number of challenges, including the continuous and non-stationary nature of data, the massive volume of information to be processed and constraints put on the computational resources. While there is a number of supervised solutions proposed for this problem in the literature, most of them assume that access to the ground truth (in form of class labels) is unlimited and such information can be instantly utilized when updating the learning system. This is far from being realistic, as one must consider the underlying cost of acquiring labels. Therefore, solutions that can reduce the requirements for ground truth in streaming scenarios are required. In this paper, we propose a novel framework for mining drifting data streams on a budget, by combining information coming from active learning and self-labeling. We introduce several strategies that can take advantage of both intelligent instance selection and semi-supervised procedures, while taking into account the potential presence of concept drift. Such a hybrid approach allows for efficient exploration and exploitation of streaming data structures within realistic labeling budgets. Since our framework works as a wrapper, it may be applied with different learning algorithms. Experimental study, carried out on a diverse set of real-world data streams with various types of concept drift, proves the usefulness of the proposed strategies when dealing with highly limited access to class labels. The presented hybrid approach is especially feasible when one cannot increase a budget for labeling or replace an inefficient classifier. We deliver a set of recommendations regarding areas of applicability for our strategies.
% \PACS{PACS code1 \and PACS code2 \and more}
% \subclass{MSC code1 \and MSC code2 \and more}
\end{abstract}

\section{Introduction}
\label{intro}

Contemporary data sources generate new information at both tremendous size and speed. Therefore, modern machine learning systems must deal not only with the volume but also velocity issues \cite{Zliobaite:2012}. Stock exchange, sensor networks, or social media are among the examples of scenarios in which new instances continuously arrive at high speed over time, creating a demand for adaptive, real-time data mining algorithms. This has led to the emergence of data streams notion and the development of a family of dedicated algorithms. New challenges needed to be addressed, such as the potentially unbounded size of the data that may quickly overflow computational resources \cite{Cano:2018}, learning schemes that are able to use both new and historic instances, as well as approaches for managing the evolving nature of data \cite{Ramirez-Gallego:2017}. A phenomenon known as concept drift is embedded in the streaming scenario, as characteristics of data may change over time \cite{Gama:2014}. 

Supervised learning has gained significant attention in data stream mining, allowing for efficient classification and prediction from non-stationary data \cite{Ditzler:2015}. However, the vast majority of research in this area assumes that class labels become available right after the incoming instance was processed. Then a label is obtained and used to update the learning system. While this is true to the overall stream mining principles, it completely neglects the issue of how to actually obtain every single label. If we would have access to a theoretical oracle that would provide us with such information every time we query for it, then what is the purpose of a classification system? In reality, a class label should be provided by a domain expert in order to maintain the highest certainty with regard to the used information. However, an expert's service is related to the cost, both in monetary and time terms \cite{Zliobaite:2015}, and thus cannot be called upon every time a new instance becomes available \cite{Abdallah:2015,Masud:2011}. This holds especially true for massive and high-speed data streams (e.g., Twitter can produce around 350 000 new tweets every 60 seconds). Therefore, methods that would allow for mining data streams on a budget are in high demand \cite{Zliobaite:2014unc}. 

\smallskip
\noindent \textbf{Summary.} We propose a novel hybrid framework for low-cost data stream mining. It is based on combining active learning and self-labeling for using available limited labels most effectively. These two strategies work in a complementary fashion, boosting their advantages while minimizing their weak sides. Active learning allows for an informative selection of instances that will be most useful for adjusting the classifier to the current state of the stream. However, each such query reduces the available budget. Self-labeling exploits discovered data structures and improves the competency of a classifier at no cost, yet offers no quality validation. The proposed hybrid approach allows active learning to explore the incoming data stream for new emerging concepts, while self-labeling offers their further exploitation without depleting the budget. We introduce 7 hybrid strategies, divided into two groups: blind and informed. The former one consists of methods that use the classifier output for deciding if a new instance should be used for active learning or self-labeling. The latter one combines the information from both the classifier and drift detector modules to better adapt decisions to the evolving nature of data streams. The analysis of the performance of these proposed hybrid algorithms sheds light on their areas of applicability, as well as allows us to identify shortcomings of state-of-the-art active learning methods.

\smallskip
\noindent \textbf{Main contributions.} This work offers the following contributions to the data stream mining domain.

\begin{itemize}
	
	\item A novel hybrid framework for mining data streams on a budget, using the combination of active and semi-supervised learning.
	
	\smallskip
	\item Two families of methods based on blind and informed approaches, leading to seven algorithms for empowering active learning with self-labeling.
	
	\smallskip
	\item Thorough experimental study on real data streams under various labeling budgets, showcasing the advantages of using hybrid solutions when available class label are scarce, especially in extremely small budget cases. 
	
	\smallskip
	\item Analysis and recommendations on areas of applicability of the proposed algorithms.
		
\end{itemize}

\section{Data stream mining}
\label{sec:dsm}

A data stream can be defined as a sequence of ordered instances arriving over time and of potentially unbounded volume. Mining data streams imposes certain specific requirements on used classifiers. Contrary to static scenarios a predefined training set is not available, as instances become available at given time intervals one by one (online case) or in form of chunks (block case). The entire stream cannot be stored in memory due to its unknown and ever-expanding size. Thus only a limited number of most recent instances can be stored, while the old ones are to be discarded to limit the computational resources being used. Characteristics of the stream may evolve over time and this must be accommodated during the continuous learning process.

More technically, a data stream is a sequence $<S_1, S_2, ..., S_n,...>$, where each element $S_j$ is a set of instances (or a single instance in case of online learning), each of them being independent and randomly generated according to a probability distribution $D_j$. If a transition between states S$_j$ $\rightarrow$ S$_{j+1}$ holds D$_j$ = D$_{j+1}$, then we deal with a stationary stream. However, in most real-life scenarios incoming data is subject to change, leading to the notion of non-stationary streams and concept drift \cite{Gama:2014}. As concept drift may affect various properties of the stream, therefore we may analyze it from various perspectives. When taking into account the influence on learned decision boundaries, one may distinguish between real and virtual concept drift. The former has an effect on posterior probabilities and may impact unconditional probability density functions. This forces the learning system to adapt to change in order not to lose competence. The latter drift does not have any effect on posterior probabilities, but only on conditional probability density functions. It may still cause difficulties for the learning system, leading to false alarms and unnecessary computational expenses due to rebuilding the classifier. Another view on the concept drift comes from the severity of ongoing changes. Sudden concept drift appears when S$_j$ is being suddenly replaced by S$_{j+1}$, where D$_j \neq$ D$_{j+1}$. Gradual concept drift is a transition phase where examples in S$_{j+1}$ are generated by a mixture of  D$_j$ and D$_{j+1}$ with their proportions continuously changing. Incremental concept drift is characterized by a smooth and slow transition between distributions, where the differences between D$_j$ and D$_{j+1}$ are not significant. Additionally, we may face recurring concept drift, in which a state from $k$-th previous iteration may suddenly reemerge D$_{j+1}$ = D$_{j-k}$, which may take place once or periodically.

Learning systems designed for data stream mining must take into account the presence of concept drift and have embedded a solution to tackle it \cite{Ditzler:2015}. Three general possible approaches include: (i) rebuilding a classifier from scratch whenever new instances become available; (ii) using a mechanism to detect change occurrence and guide the rebuilding process more effectively; and (iii) using an adaptive classifier that will adjust automatically to the current state of the stream. The first approach is impractical, due to prohibitive costs connected with continuous model deleting and rebuilding. Therefore, the two remaining directions are commonly utilized in this domain. Concept drift detectors are external tools that allow for monitoring the characteristics of a stream, in order to anticipate and detect the change point \cite{Korycki:2021}. Adaptive classifiers are based on either sliding windows \cite{Wozniak:2011} or on online learners \cite{Tu:2018}. Finally, ensemble solutions are popularly used for mining drifting data streams \cite{Krawczyk:2017}.

\section{Labeling constraints in data streams}
\label{sec:lcs}

While supervised learning from drifting data streams has gained significant attention in recent years, the vast majority of developed algorithms assume working in \textit{test-then-train} mode. This means that each new instance (or chunk) is first used to test the current classifier and then utilized for the updating procedure. This scheme easily translates into real-life scenarios, when we use the model at hand to predict labels for new data and then try to conduct the learner update. However, many of these algorithms work under the assumption that true class labels become available the moment after we obtain predictions from the model. While such a set-up is convenient for an experimental evaluation of a classifier in the controlled environment, it completely fails to capture the real-life situation of dealing with limited ground truth availability \cite{Lughofer:2016}. Therefore, many supervised classifiers, while delivering excellent performance on benchmark data, cannot be directly used in practical applications \cite{Krawczyk:2017}.

This is caused by the fact that obtaining a true class label for a new instance is far from being a trivial task. If we would have access to a theoretical oracle that would provide us with such information every time a new instance becomes available, then there is no need to have any classification procedures. In most real-life applications, a domain expert is required to analyze a given instance and label it. While one may theorize that a company developing a specific data stream mining system should have such an expert at its disposal, we cannot forget the costs connected with such a procedure. This can be viewed as monetary costs, as an expert would require a payment for sharing his knowledge, as well as a time cost, as an expert needs to spend some time analyzing each instance. Therefore, in a real-life scenario, neither a constant label query is possible (as a given company would quickly use up its budget), nor instant label availability \cite{Grzenda:2020}. Even if these factors, for various reasons, play a less important role, the human throughput must also be considered. A given expert cannot work non-stop and will have limited responsiveness per given time unit. Thus, in cases of massive and high-speed data streams assumption of continuous label availability cannot hold.  

There are specific application areas, where true class labels can be obtained at no cost. Let us consider stock exchange or weather prediction. Here one may observe the current state of the environment in order to verify the prediction. However, one still cannot assume that the ground truth will become instantly available. Usually, such predictions are being made forward into the future, whether we consider hours, days or months. Therefore, a given amount of time must pass before a ground truth can be observed. Although here we do not have the problem of associated cost, we still need to deal with label latency \cite{Plasse:2016}.

This has led to works on learning from data streams under limited access to class labels. Most popular approaches include active learning solutions that allow selecting only a limited number of instances for labeling \cite{Krawczyk:2019}. They are usually selected to offer new information to the classifier, instead of reinforcing old concepts \cite{Martins:2020}. Although there is a good amount of research on active learning for static scenarios \cite{Pupo:2018}, there exist but a few solutions that take into account the drifting and evolving nature of streams \cite{Korycki:2019ali,Zliobaite:2014unc}. Another branch of works focuses on semi-supervised learning \cite{Masud:2011}, usually using clustering-based solutions \cite{Li:2012}. The most common assumption here is that with every time interval a subset of instances arrives as labeled and one may use them to guide the learning process. This is fundamentally different from active learning, as it concentrates on how to make use of unlabeled instances, instead of selecting the proper ones for labeling. There also exist algorithms working under the assumption that only initial data is being labeled and no further ground truth access is to be expected \cite{Dyer:2014}.

As we deal with limited access to class labels, any techniques that may allow us to better exploit the given budget and take advantage of unlabeled instances without paying the cost for querying them are highly attractive for data stream mining \cite{Zhu:2020}. Therefore, we focus our attention on combining active and semi-supervised \cite{Triguero:2015} learning paradigms. Many hybrid frameworks have been proposed for batch mode, offline settings and successfully applied to different domains. Combining self-labeling based on uncertainty with active learning strategies is just one possible approach, where the least certain instances are chosen to be queried and the most certain are labeled by a classifier on its own. Graph-based methods build a similarity graph to determine which samples are the most informative and should be queried. The labeled objects are then used to propagate their classes to the closest neighbors in the graph.

The area of hybrid frameworks for evolving data streams mining is a relatively recent idea and yet weakly explored. As far as we know, while there is a lot of pure active learning \cite{Liu:2021,Shan:2019} or semi-supervised learning \cite{Pratama:2021,Zheng:2021} methods for streaming data, very few hybrid solutions have been proposed. One of them uses variance reduction on data chunks as an active learning strategy and the same instance selection method as a self-labeling step \cite{Kholghi:2011var}. Another solution is based on the incremental graph building algorithm -- Adaptive Incremental Neural Gas (AING) -- which provides information needed to decide if a sample should be queried or if a classifier can teach itself with it \cite{Bouguelia:2013aing}. An interesting framework was proposed in \cite{Goldberg:2011oasis}. It is described as a highly scalable, parallelizable and optimized online Bayesian framework using sequential Monte Carlo and so-called gap assumption. In our previous work, we have proposed the first study on a combination of active and self-labeling solutions for data streams \cite{Korycki:2018}. However, the algorithm described there used a fixed threshold and therefore was unable to efficiently adapt to the presence of concept drift. Addressing this limitation is the cornerstone of the study proposed in this manuscript.

Due to the lack of comprehensive works on hybrid frameworks for data streams, we investigate a framework that offers a combination of active learning with different self-labeling strategies based on uncertainty thresholding in the presence of concept drifts and under labeling constraints. While most of the presented solutions take for granted fixed thresholds, which are core elements of proper decisions, we focus on changing the self-labeling threshold adaptively to an incoming data stream.

\section{Combining active learning with self-labeling for drifting data streams}
\label{sec:als}

The reduction of the labeling cost can be intensified using different methods concurrently. Both active learning and semi-supervised learning algorithms aim to enhance the learning process when few labeled instances are available and there is a vast number of unlabeled samples. However, they focus on generally disjunctive facets of data utilization. The former group of methods uses limited supervision for selecting only the most informative samples to be labeled, while the latter exploits already gained knowledge to adjust a model more properly in an unsupervised manner. Therefore, active learning can be interpreted as an exploration process, since it \textit{asks for the unknown}, and semi-supervised learning as an exploitation step -- it \textit{exploits the known} \cite{Settles:2010al}. 

\begin{figure}[h]
	\centering
	\includegraphics[scale=0.5]{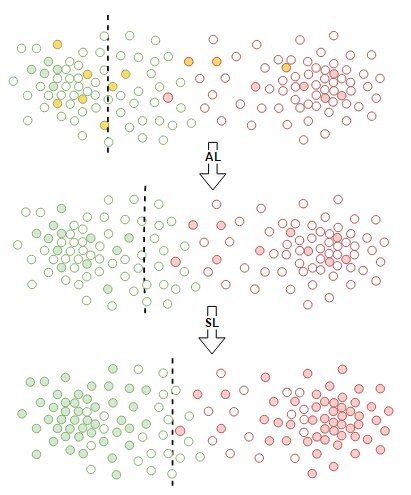}
	\caption{The idea of combining active learning and self-labeling. (Top) The original linear model; (middle) the decision boundary updated after utilizing labeled instances selected by the active learning query; (bottom) self-labeling improving the decision boundary at no additional labeling cost, due to smoothness assumptions being fulfilled.}
	\label{fig:hybrid_idea}
\end{figure}

The idea behind such a combination is depicted in Fig. ~\ref{fig:hybrid_idea}. It depicts a situation in which the decision boundary is successively moved away from the green cluster's center. At first, the labeled instances selected by the active learning algorithm delineate a general decision boundary. Since this dataset fulfills the smoothness assumption \cite{Chapelle:2010book}, during the self-labeling step it is more likely that objects will be taken from the dense cluster on the left, so the classification will be correct and the border will be moved in the right direction at no additional cost. This can be explained by the fact that active learning itself, especially on low budgets, will not be able to exhaustively sample the properties of each class. Therefore, we may deal with the issue of underfitting, where we do not have enough labeled instances to correctly estimate the decision boundary. Self-labeling allows for correcting this problem by using the small number of labeled instances to enrich the training set for each of the classes. We find it intuitive to consider some hybrid frameworks which offer a hybridization of both techniques to reduce the overall amount of labeled instances needed for an effective model construction under drifting data streams.

\subsection{Framework}
\label{sec:framework}

In this section, we present our generic online hybrid framework for active and semi-supervised hybridization. The framework consists of four flexible components.

\begin{itemize}

\item \textbf{Adaptive classifier:} a backbone of our framework that can be realized by any learning algorithm capable of incremental/online processing of instances. 

\smallskip
\item \textbf{Concept drift detector:} a module for monitoring the changes in the stream and informing our framework when it is necessary to take a drifting nature of a distribution into account.

\smallskip
\item \textbf{Active learning strategy:} a module for selecting the most valuable instances for the label query process that allows us to obtain labeled instances under the given budget constraints. 

\smallskip
\item \textbf{Self-labeling strategy:} a module allowing for handling the underfitting of current concepts that uses instances acquired by active learning to increase the size of labeled training set at no additional cost imposed on the budget. 

\end{itemize}

\noindent \textbf{Adaptive classifier} is being updated with incoming data samples. It is important that the algorithm should be able to work in an online setting since the whole framework is designed to do so. Another constraint is that a classifier should return a probabilistic outcome or values which can be approximately treated as the posterior probability. Some of them are Na{\"i}ve Bayes or k-NN algorithm, to name a few. The \textit{state-of-art} classifier, which is able to classify data streams efficiently and is responsive to a non-stationary distribution, is Adaptive Hoeffding Tree (AHT) \cite{Bifet:2009hat}. It is a modification of the incremental decision tree called Very Fast Decision Tree (VFDT) \cite{Domingos:2000vfdt} which uses the Hoeffding bound to expand its structure. AHT applies adaptive windows on every node to adjust local statistics independently and replace outdated nodes, so it is able to adapt to concept drifts. There are a few other variants of Hoeffding Tree - one of them is Randomized Hoeffding Tree (RHT) \cite{MOA:2018} that introduces additional randomization to the learning process. More complex classifiers like online ensembles can also be used in our framework. For example, there is Recurring Concept Drift framework (RCD), which handles drifts by storing a separate base learner for each of distinguished concepts \cite{GoncAlvesJr:2013}, or Accuracy Weighted Ensemble (AWE), which maintains a committee of models that are weighted by their accuracy on incoming data \cite{Wang:2003}. We find our hybrid framework for learning on a budget highly flexible, so it can be adjusted to many real-life data stream mining scenarios.

\medskip
\noindent \textbf{Concept drift detector} is responsible for providing indications of changes in a data stream. The information can be used by different strategies to adjust their parameters and boost the overall performance. The most popular drift detectors are Drift Detection Method (DDM) \cite{Gama:2004ddm} and Early Drift Detection Method EDDM \cite{Garcia:2006eddm}, which will be described later.

\medskip
\noindent \textbf{Active learning strategy} can be realized via any online strategy from the group of stream-based selective sampling methods \cite{Settles:2010al}. In our work, we incorporate an uncertainty sampling strategy \cite{Aggarwal:2014al} which makes its decision based on some uncertainty measures -- usually, they use a maximum a posteriori probability or an entropy. The measure is compared with a threshold. If it is low enough, a sample should be queried. For the former measure the inequality is given as follows:
\begin{equation}\label{eq:maxpost}
p(\hat{y}|X) = \max_y(p(y|X)) \leq \theta,
\end{equation}

\noindent where $X$ is a vector of an instance's features, $p(y|X)$ is the conditional class probability and $\theta$ is the threshold. The lower the threshold, the less certain samples are chosen to be queried. Therefore, lower values are preferred. However, one should notice that too strict conditions may favor only those objects which are close to a decision boundary, so changes that occur far from the boundary may never be detected

The issue of determining the appropriate threshold for uncertainty sampling strategies has been comprehensively studied in \cite{Zliobaite:2014unc}. We construct our active learning module based on the results from this work. One of the proposed reliable strategies is Variable Uncertainty Strategy with Randomization (\texttt{RandVar}). In Alg. \ref{alg:randvar}, each incoming instance $X$ is forwarded to a classifier $L$ which returns posterior probabilities for the sample. The uncertainty measure is determined through the maximum a posterior rule. There is a threshold $\theta$ that is used for decision making. These are usual steps. The strategy introduces two additional operations.

\begin{algorithm}[h]
	\KwData{incoming instance $X$, trained classifier $L$,
		threshold adjustment step $s \in (0,1 \rangle$, variance of the threshold randomization $\sigma$}
	\KwResult{\textit{labeling} $\in \{true,false\}$}
	\KwInit{$\theta \gets 1$, store the latest value during operation}
	$\hat{y} \gets \argmax_y p(y|X)$, where $y \in \{1,...,c\}$\;
	$\theta_r \gets \theta \times \eta$, where $\eta \in  N(1,\sigma)$ is a random multiplier\;
	
	\uIf{$p(\hat{y}|X) \leq \theta_r$}{
		decrease the uncertainty region $\theta \leftarrow \theta(1-s)$\;
		\KwRet{ labeling $\gets true$}
	}\Else{
		increase the uncertainty region $\theta \leftarrow \theta(1+s)$\;
		\KwRet{ labeling $\gets false$}
	}
	\caption{Variable Uncertainty Strategy with Randomization}
	\label{alg:randvar}
\end{algorithm}

Firstly, it modifies the threshold value with an adjustment step $s$. The idea is to lower the value when a concept drift occurs and the classifier returns more uncertain labels more frequently; and to increase it when there is a stable period and the model is more sure about its decisions. This step leads to a more balanced budget spending, so the labeling is more uniformly distributed over time. Secondly, the threshold is randomized by multiplying it by a variable drawn from the Gaussian distribution with mean $m=1$ and predefined variance $\sigma$. This step ensures that a decision space is sampled more uniformly since the instances which are far from the decision boundary are more likely to be queried. We find the strategy universal and according to the presented results reliable, so we choose it as our base active learning strategy. We do not perform any further investigation on the matter. Instead, we focus on selecting the appropriate self-labeling strategy which will cooperate well in the hybrid framework.

\medskip
\noindent \textbf{Self-labeling strategy} complements the selected active learning strategy, acting as our semi-supervised learning approach. We assume that unsupervised learning based on the knowledge provided by the queries will lead to more exhaustive and accurate adaptation on the same budget. In addition, we preconceive that the active learning strategy will guide self-labeling properly, thus reducing the risk of error amplification. This module should also work in an online setting. Relating to the previous method, we propose different strategies for changing the self-labeling threshold respectively to an incoming data stream. We investigate self-reliant approaches, as well as, using a feedback from the active learning strategy and drift detectors. To the best of our knowledge, such research has not been done before. 

\begin{algorithm}[h]
	\KwData{labeling budget $B$, \texttt{QueryStrategy}, \texttt{SelflabelingStrategy}, \texttt{DriftDetector}}
	\KwResult{classifier $L$ at every iteration}
	\KwInit{$\hat{b} \gets 0$}
	\Repeat{stream ends}{
		receive incoming instance $X$\;
		get class label $\hat{y}$ predicted by classifier $L$\;
		\uIf{$\hat{b} < B$ {\normalfont and} \upshape\texttt{QueryStrategy} $(X) = true$}{
			request the true label $y$ of instance $X$\;
			update labeling expenses $\hat{b}$\;
			update drift indications using \texttt{DriftDetector} $(y, \hat{y})$\;
			update classifier $L$ with $(X,y)$\;
		}\ElseIf{\upshape\texttt{SelflabelingStrategy} $(X, feedback) = true$}{
			update classifier $L$ with $(X,\hat{y})$\;	
		}	
	}
	\caption{The generic hybrid algorithm for combining active learning and self-labeling}
	\label{alg:framework}
\end{algorithm}

The proposed generic algorithm is presented in pseudo-code form as Alg.~\ref{alg:framework}. For each incoming sample, the actual budget spending $\hat{b}$ is being checked. For potentially infinite streams it can be estimated as a ratio of already labeled instances to all received samples. If the value does not exceed a given budget $B$ and a response from the active learning strategy (\texttt{QueryStrategy}) is positive, the sample is queried. After receiving a label, the expenses, drift indications and classifier are updated. If the query strategy rejects the sample, it still can be subject to the self-labeling approach (\texttt{SelflabelingStrategy}). This depends on the response of the second strategy we add to the classification flow. The method may also be optionally supplied with a $feedback$ from the active learning strategy or the drift detector. It is worth saying that the chosen strategies and their measures are, in fact, complementary. Very uncertain data samples will be forwarded to an expert and, obviously, they will not be used for the self-training. On the other hand, the algorithm will discard the objects for which it will be very confident and will use them for the training without additional supervision. We put the active learning step before the self-labeling, because we assume that the latter requires sufficient guidelines from the former, thus it is better to spend the whole available budget. Proposed self-labeling strategies for the hybrid online approach are described in the next sections.

\subsection{Blind self-labeling strategies}
\label{sec:blind}

The first group of strategies are blind approaches \cite{Gama:2014}. These methods do not use any explicit information about changes that occur in a data stream. They make some assumptions and use heuristics to handle data evolution. Classifiers, which are guided by them, adapt gradually to concept drifts, slowly forgetting old concepts and learning new ones. It takes time, as there is no clear and direct indication of a change. Some blind strategies for active learning were proposed in the mentioned work \cite{Zliobaite:2014unc}. Another one is, for example, the adaptation method used in VFDT that incorporates a sliding window technique for its statistics \cite{Domingos:2000vfdt}.

\subsubsection{Fixed}
\label{sec:fixed}

The first and the simplest blind strategy we use in our hybrid framework is based on a fixed threshold -- we call it \texttt{Fixed} strategy. To determine a confidence measure, which is compared with the self-labeling threshold $\gamma$, the maximum a posteriori rule is used, since it returns a value that is chosen by a model to classify an instance:
\begin{equation}
p(\hat{y}|X) = \max_y(p(y|X)) \geq \gamma.
\end{equation}

\noindent The inequality is opposite to the active learning condition. It prefers those samples to which a classifier displays the highest certainty, so there is a higher possibility that a classifier's own decision will be correct. This strategy is the most popular one in semi-supervised streaming frameworks discussed before, as the research on the self-labeling module is usually very limited.

\subsubsection{Uniform and Randomized Uniform}
\label{sec:uniform}

The main weakness of the previous strategy is the need to predefine the threshold value. During a concept drift reliability of a classifier may significantly change, so a distribution of posterior probabilities will also be different -- there will be more lower values than during a stable period. It is not obvious which values and when can be denoted as \textit{high enough} or even that they should be actually high. 

Our two following strategies are based on the methods presented in \cite{Zliobaite:2014unc}. The first one is \texttt{Uni}. It incorporates the idea of balancing a budget spending over time (Alg. \ref{alg:randvar}). Since there is no budget for self-labeling, we simply aim to ensure that the process will be uniformly frequent, regardless of a stream's state.

\medskip
\begin{algorithm}[]
	\KwData{incoming instance $X$, trained classifier $L$,
		threshold adjustment step $s \in (0,1 \rangle$}
	\KwResult{\textit{self-labeling} $\in \{true,false\}$}
	\KwInit{$\gamma \gets 1$, store
		the latest value during operation}
	$\hat{y} \gets \argmax_y p(y|X)$, where $y \in \{1,...,c\}$\;
	
	\uIf{$p(\hat{y}|X) \geq \gamma$}{
		decrease the confidence region $\gamma \leftarrow \gamma(1+s)$\;
		\KwRet{ self-labeling $\gets true$}
	}\Else{
		increase the confidence region $\gamma \leftarrow \gamma(1-s)$\;
		\KwRet{ self-labeling $\gets false$}
	}
	\caption{\texttt{Uni} self-labeling strategy}
	\label{alg:uniform}
\end{algorithm}

We achieve this by increasing the threshold when the classifier $L$ becomes more confident, so we prevent it from being overfitted. On the other hand, we decrease the threshold when some changes occur and decisions are less confident, therefore, the model should be sufficiently supplied with additional labels and adjust itself to changes faster. 

The second one is Randomized Uniform (\texttt{RandUni}) strategy. In fact, it is \texttt{RandVar} strategy (Alg. \ref{alg:randvar}) with increasing and decreasing the threshold as in the previous algorithm. We assume that the randomization step may assure some helpful diversification, especially after abrupt changes, when $\gamma$ is usually far from a more stable value.

\subsubsection{Inverted Uncertainty}
\label{sec:invunc}

The last blind strategy is Inverted Uncertainty (\texttt{InvUnc}). In this approach, we use the active learning threshold $\theta$ from \texttt{RandVar} algorithm to control the self-labeling threshold $\gamma$ using a simple transformation\footnote{One may call it an informed strategy, however, since the self-labeling threshold is controlled by the value determined with a blind strategy, we consider \texttt{InvUnc} also a blind method.}. We assume that $\gamma$ should be higher during a concept drift and lower for stable periods. It is motivated by an assumption that when a concept is changing and a model is insufficiently adapted, there are no formed internal class structures to be exploited, so only the most certain decisions should be made. On the other hand, when a concept is stable these structures are more likely to be present, therefore, we  should intensify self-labeling in order to be able to exploit them. In \texttt{RandVar} the threshold is higher during stable periods and lower when drifts occur. The self-labeling condition is then given as follows:
\begin{equation}
p(\hat{y}|X) \geq 1 - \theta + \frac{1}{c},
\end{equation}

\noindent where $p(\hat{y}|X)$ is a maximum a posteriori probability and $c$ is a number of possible classes. $1 - \theta$ factor is self-explanatory. That is why we call the method Inverted Uncertainty. We add $1/c$ factor, due to the fact that $p(\hat{y}|X)$ is never lower than it, so regardless of small disturbances introduced by the randomization step for small values of $s$ (which should be used \cite{Zliobaite:2014unc}), the threshold $\theta$ is in range $\langle \frac{1}{c},1 \rangle$. We want $\gamma$ to be in the same range.

\subsection{Informed self-labeling strategies}
\label{sec:informed}
To improve the responsiveness of the adaptation process dedicated change detectors can be introduced. Those algorithms are directly oriented on indicating changes, thus they provide accurate information about stream evolution. Strategies that use such indicators are called informed \cite{Gama:2014}. The simplest drift detector can be based on a classification error rate. The two most popular algorithms that have been already mentioned are DDM and EDDM. We use them as a base of our next three self-labeling strategies, in order to improve their adaptation capabilities in the presence of concept drift.

\subsubsection{Continuous DDM}
\label{sec:cddm}
In this approach, we use DDM indications to create continuous control over the self-labeling threshold. We call the method Continuous DDM (\texttt{cDDM}). The standard DDM is designed to indicate three discrete states of a data stream: \textit{stable}, \textit{warning} and \textit{change}. The core assumption of this method is that an error rate $p$ of a classifier should be approximately constant and low when a concept is stable. When a drift occurs, the error should be significantly higher than the value registered for the static distribution, since the classifier has not adapted to a new concept yet. Therefore, changes can be detected by tracking the actual error rate $p$ along with its standard deviation $s$ and comparing it with the registered error for the stable period. The algorithm makes decisions based on the condition:
\begin{equation}
p + s > p_{min} + \alpha s_{min},
\end{equation}

\noindent where $p_{min}$ and $s_{min}$ are the mean error and its standard deviation registered for a stable concept after at least 30 samples. The $\alpha$ parameter is used to determine thresholds for \textit{warning} ($\alpha = 2$, the confidence interval is 95\%) and \textit{change} ($\alpha = 3$, the confidence interval is 99\%) states \cite{Gama:2004ddm}. Such discrete DDM can be used to reset a classifier and itself when a drift occurs \cite{Zliobaite:2014unc}.

Since we want to control the self-labeling threshold continuously and respectively to the DDM indications, we take the whole algorithm as it is, excluding the classifier reset, and we simply extract the tracked, continuous error measure $\epsilon = p + s$. We use this value analogously to the \texttt{InvUnc} idea (Sec. \ref{sec:invunc}) -- the threshold should be higher during a concept drift and lower during a stable period. Decisions are based on the following condition:
\begin{equation}
p(\hat{y}|X) \geq \tanh{2(\epsilon + \frac{1}{c})}.
\end{equation}

\noindent We add $1/c$ to additionally penalize a situation when a classifier simply guesses labels for $\epsilon = 1 - 1/c$. For example when $\epsilon = 0.5$ for two classes, then $\epsilon + \frac{1}{c} = 1$, which is the maximum possible posterior probability returned by the model. Since the error can be higher than $1-1/c$ we have to normalize the overall value -- we use $\tanh2x$ for this, which is approximately correct, so $\gamma$ is never higher than 1. At the same time, the lowest threshold $\gamma$ is always higher than $1/c$, therefore, the condition in this strategy is a bit stricter than in the others. It can be justified since for posterior probabilities close to $1/c$ a classifier is very uncertain about its decisions.

\subsubsection{Continuous EDDM}
\label{sec:ceddm}
Another drift detector -- EDDM -- presents a slightly different approach. We use it for our second informed strategy called Continuous EDDM (\texttt{cEDDM}). Instead of considering the mean error, the algorithm calculates an average distance $p'$ between two misclassified objects and also its standard deviation $s'$. When a concept is stable, the average distance increases due to a model adaptation process. When the concept starts to change, gaps between two mistakes become shorter. The algorithm compares the current mean distance with the maximum registered:
\begin{equation}
\label{eq:eddm}
\frac{p' + 2s'}{p'_{max} + 2s'_{max}} < \beta,
\end{equation}

\noindent where $p'_{max}$ and $s'_{max}$ are registered statistics for a stable period after at least 30 errors. $\beta$ is a threshold for a similarity between two error distance distributions. It is empirically recommended to set $\beta = 0.95$ for \textit{warning} and $\beta = 0.9$ for \textit{change} \cite{Garcia:2006eddm}. The standard EDDM is used just like DDM, including resetting.

Again, to control self-labeling in a continuous way, we remove the classifier reset step and extract the similarity measure $\zeta \frac{p' + 2s'}{p'_{max} + 2s'_{max}}$, as the reference value. One should notice that $\zeta \in \langle 0.9, 1 \rangle$, because when a drift is detected for $\zeta = 0.9$ the EDDM is reset and $\zeta$ stays unchanged until the next update is possible. The idea of the control is the same as for \texttt{InvUnc} and \texttt{cDDM}. The condition that is used to check if we should use an instance in the self-labeling process is given as follows:
\begin{equation}
p(\hat{y}|X) \geq f(\zeta),
\end{equation}

\noindent where $f(\zeta)$ is any decreasing function defined as:
\begin{equation}
f(\zeta) : \{\zeta \in \langle 0.9, 1 \rangle, f(0.9) = 1, f(1) = \frac{1}{c}\}.
\end{equation}

\noindent In our case we choose a simple linear function which fulfills the above requirements.

\subsubsection{Windowed Error}
\label{sec:window}
The last strategy is Windowed Error (\texttt{WinErr}). This method is, indeed, a sliding window for the mean error. Instead of resetting drift indications $p$ and $s$, like in \texttt{cDDM}, we track the strictly continuous error that changes dynamically within the window. We use the same self-labeling condition as in \texttt{cDDM}.

\medskip
It is worth noting that a limited labeling budget may have a significant influence on indications generated by drift detectors. Due to the lack of a sufficient amount of error indicators (coming with labels), internal estimators used in the detectors will be inaccurate, impeding the process of drift detection. It may be a serious problem in the case of discrete (binary) drift detection when one wants to use these indications to reset a model and retrain it from scratch since if there is not enough information the detectors may never be triggered or they may act at random. However, in our work, we do not use drift detectors in such a way. Instead of working in the retraining mode, we continuously update our strategies in an informed manner using the error (DDM) or distance between errors (EDDM) as a continuous input to our self-labeling strategies. By doing this we alleviate the mentioned problem since we do not have to rely on rare, binary and unreliable drift detections. Obviously, we may still be forced to use imperfect estimations, especially for very low budgets, but it is something we have to accept, due to the assumption of strictly limited access to ground truth. There is very little that can be done without more labels.

Still, the used drift detectors could be further investigated in the given settings and it is possible that some improvements can be done in order to improve their performance while working on a budget. However, since that would require a broader in-depth statistical analysis in various scenarios we find it beyond the scope of this work.

\section{Experimental study}
\label{sec:exp}
In this section, we describe our experiments conducted to prove the validity of introduced strategies for active learning and self-labeling hybridization. Firstly, we present data streams that are used in the evaluation process. Next, we briefly delineate the problem of measuring performance in a streaming data environment and describe a chosen set-up. Finally, obtained results with a commentary and conclusions are presented.

\subsection{Set-up}
\label{sec:dat}

\noindent \textbf{Data streams.} To examine the adaptation capabilities of the presented algorithms we used a set of drifting data streams. We attempted to explore our strategies in the context of a variety of data streams and concept drifts. Therefore, we utilized real streams from different domains and with diversified properties. The real data streams allow for conducting relatively reliable tests of adaptive methods in real-world environments. They are characterized by mixed types of drifts that are coming from the underlying nature of datasets, thus leading to more realistic learning difficulties. Details of used real streams are given in Tab.~\ref{tab:real}.

\begin{table}[h]
	\caption{Summary of the used real data streams.}
	\centering
	\begin{tabular}[H]{lccccc}	
		\toprule
		Name & Inst & Attr & Cls & Type  \\ 
		\midrule
		Sensor & 2 219 803 & 5 & 54 & numeric \\
		Cover & 581 012 & 54 & 7 & numeric, nominal \\ 	 		
		Spam & 9324 & 499 & 2 & textual \\	
		Power & 29 928 & 2 & 24 & numeric \\
		Usenet & 5931 & 658 & 2 & textual \\
		Gas & 13 910 & 128 & 6 & numeric \\
		Poker & 829 201 & 10 & 10 & nominal \\
		Elec & 45 312 & 8 & 2 & numeric, nominal \\
		\bottomrule
	\end{tabular}
	\label{tab:real}
\end{table}

\medskip
\noindent \textbf{Evaluation methods.} Evaluation in streaming data environments enforces a different approach than those which are used in standard batch mode scenarios. Such reliable methods like k-fold cross validation may turn out to be impractical due to its time-consuming nature. Furthermore, other techniques which are dedicated to data streams, like holdout \cite{Lemaire:2015holdout}, may be inappropriate for streams with concept drifts. Simpler error-estimation procedures that cope with dynamic online settings are possible. One method, which can be successfully used for the considered evaluation, is prequential evaluation $p_e$ with sliding window for the most recent examples \cite{Bifet:2015preq}:
\begin{equation}
p_e = \frac{1}{\omega}\sum_{k=t-w+1}^{t}L(y,y^*),
\end{equation}

\noindent where $\omega$ is the window size and $L(y,y^*)$ is a chosen loss function for predicted outputs $y$ and true labels $y^*$. Using this approach, each sample is firstly utilized for testing and later for learning. The chosen accuracy measure is recalculated instance by instance. It provides a relatively good measurement sensitivity, but it highly depends on the optimal window size. We chose this approach with $\omega=1000$ to estimate an average error within the window for time series measurements and, in addition, to calculate a global average for a whole stream.

\medskip
\noindent \textbf{Examined strategies.} In our experiments, we examined the proposed strategies: \texttt{Fixed}, \texttt{Uni}, \texttt{RandUni}, \texttt{InvUnc}, \texttt{cDDM}, \texttt{cEDDM} and \texttt{WinErr} on the presented real data streams, using the chosen evaluation method. As a baseline, we selected a pure active learning strategy \texttt{RandVar} introduced in Sec. \ref{sec:framework}. The variable threshold was set to $s = 0.01$ and the standard deviation of randomization to $\sigma = 1$. We call the strategy \texttt{ALRV} in the experiments. In addition we include results for random selection (\texttt{ALR}) \cite{Zliobaite:2014unc} and sampling (\texttt{ALS}) \cite{Bianchi:2006}. For the \texttt{Fixed} strategy we empirically found out that very high threshold values are preferable on average, therefore we set $\gamma=0.95$. Parameters of \texttt{Uni} and \texttt{RandUni} were the same like in \texttt{ALRV}. The internal configuration of drift detectors for \texttt{cDDM} and \texttt{cEDDM} was set to default, as stated in \cite{Gama:2004ddm,Garcia:2006eddm}, so a minimum number of registered samples was $n=30$ and for errors it was also $n_e=30$. The width of \texttt{WinErr}'s window was set to $w = 100$, based on empirical observations that smaller windows are more reactive and accurate when a rate of information is low (a limited labeling budget), which is somehow indicated in the literature \cite{Bifet:2015preq}.

\medskip
\noindent \textbf{Labeling budgets.} We compared effectiveness of these strategies for different but generally low and very low budgets $B = \{1\%,5\%,10\%,20\%,50\%\}$. As base learners, we chose different classifiers to show that the presented framework is indeed generic. We picked two single classifiers and two ensembles that have been already mentioned in Sec. \ref{sec:framework}. These are two different Hoeffding Trees -- AHT, RHT and two ensembles -- RCD with Na{\"i}ve Bayes as base learners and AWE with perceptrons. They were tested with all the presented strategies\footnote{The experiment along with the framework implementation can be downloaded from: \textit{github.com/hybridresults/alsl20}.}.

\subsection{Results}
\label{sec:realres}

The results are presented in two forms: tables and performance series. For the former, the overall average accuracy is given. The best results are in bold. In addition, cells with scores higher than the best active learning strategy are in the green color. For the graphs presenting performance series, the accuracy within the sliding window is included. These are the best AL and self-labeling results obtained for each presented case.

\begin{table*}[]
	\caption{AHT -- the average accuracy for real streams given a budget.}
	\centering
	
	\scalebox{0.54}{
		\subfloat{\begin{tabular}[H]{lccccccc}	
				\midrule
				Stream & Strategy & $B=1\%$ & $B=5\%$ & $B=10\%$ & $B=20\%$ & $B=50\%$ \\
				\midrule
				\multirow{9}{*}{Sensor} 
				& \texttt{ALRV} & 10.80\% & 13.17\% & 17.51\% & 35.37\% & 55.69\% \\ 
				& \texttt{ALR} & 11.03\% & 15.85\% & 30.05\% & 39.44\% & 57.54\% \\
				
				& \texttt{ALS} & 7.23\% & 17.35\% & 26.60\% & 32.81\% & 56.90\% \\
				& \texttt{Fixed} & \textcolor{g}{\textbf{13.42\%}} & \textcolor{g}{24.74\%} & \textcolor{g}{44.20\%} & \textcolor{g}{\textbf{52.32\%}} & \textcolor{g}{\textbf{69.57\%}} \\
				& \texttt{Uni} & 8.23\% & \textcolor{g}{18.53\%} & 24.23\% & 26.82\% & 53.74\% \\	
				& \texttt{RandUni} & 7.64\% & 9.80\% & 12.57\% & 24.11\% & 53.76\% \\	
				& \texttt{InvUnc} & 8.53\% & 10.92\% & 13.45\% & 19.90\% & 48.02\% \\	
				& \texttt{cDDM} & 8.69\% & \textcolor{g}{18.75\%} & 23.85\% & 32.04\% & 52.33\% \\	
				& \texttt{cEDDM} & 8.62\% & 12.95\% & 14.70\% & 24.21\% & 50.52\% \\
				& \texttt{WinErr} & 9.59\% & \textcolor{g}{\textbf{27.95\%}} & \textcolor{g}{\textbf{45.88\%}} & \textcolor{g}{50.98\%} & \textcolor{g}{64.83\%} \\
				\midrule
				\multirow{9}{*}{Cover} 
				& \texttt{ALRV} & 57.85\% & 65.23\% & 73.05\% & \textbf{76.97\%} & 80.01\% \\ 
				& \texttt{ALR} & 58.74\% & 64.63\% & 72.93\% & 75.82\% & 79.51\% \\
				
				& \texttt{ALS} & 58.79\% & 64.71\% & \textbf{73.52\%} & 76.29\% & 79.45\% \\
				& \texttt{Fixed} & 22.85\% & \textcolor{g}{\textbf{68.15\%}} & 70.93\% & 74.72\% & \textcolor{g}{\textbf{81.69\%}} \\
				& \texttt{Uni} & \textcolor{g}{58.18\%} & 62.18\% & 63.02\% & 65.04\% & 78.53\% \\	
				& \texttt{RandUni} & \textcolor{g}{59.31\%} & 56.46\% & 60.45\% & 56.69\% & 77.68\% \\	
				& \texttt{InvUnc} & \textcolor{g}{60.26\%} & 56.38\% & 60.88\% & 59.20\% & 74.10\% \\	
				& \texttt{cDDM} & 58.53\% & 62.36\% & 62.55\% & 64.19\% & 75.87\% \\	
				& \texttt{cEDDM} & \textcolor{g}{\textbf{62.50\%}} & \textcolor{g}{65.38\%} & 65.50\% & 72.37\% & 79.87\% \\
				& \texttt{WinErr} & 25.04\% & 52.24\% & 66.45\% & 69.43\% & 78.57\% \\
				\midrule
				\multirow{9}{*}{Spam} 
				& \texttt{ALRV} & 27.87\% & 54.74\% & 74.87\% & \textbf{82.01\%} & \textbf{88.52\%} \\ 
				& \texttt{ALR} & 27.63\% & 54.47\% & 67.28\% & 81.08\% & 88.22\% \\
				
				& \texttt{ALS} & 27.75\% & 53.75\% & 74.39\% & 81.98\% & 88.26\% \\
				& \texttt{Fixed} & \textcolor{g}{37.19\%} & \textcolor{g}{64.68\%} & 56.79\% & 77.99\% & 86.00\% \\
				& \texttt{Uni} & 27.61\% & 33.19\% & 67.70\% & 71.49\% & 85.80\% \\	
				& \texttt{RandUni} & \textcolor{g}{57.14\%} & 29.72\% & 69.49\% & 73.95\% & 85.29\% \\	
				& \texttt{InvUnc} & \textcolor{g}{\textbf{87.31\%}} & \textcolor{g}{\textbf{87.10\%}} & 53.59\% & 79.67\% & 85.60\% \\	
				& \texttt{cDDM} & \textcolor{g}{77.77\%} & \textcolor{g}{79.40\%} & \textcolor{g}{77.25\%} & 75.12\% & 85.44\% \\	
				& \texttt{cEDDM} & \textcolor{g}{63.00\%} & \textcolor{g}{70.91\%} & \textcolor{g}{\textbf{84.18\%}} & 73.75\% & 86.04\% \\
				& \texttt{WinErr} & 25.74\% & \textcolor{g}{57.04\%} & 60.59\% & 79.91\% & 86.71\% \\
				\midrule
				\multirow{9}{*}{Power} 
				& \texttt{ALRV} & 13.83\% & 13.73\% & 14.86\% & 15.65\% & 15.50\% \\ 
				& \texttt{ALR} & 13.89\% & 13.98\% & 14.05\% & 14.64\% & 15.41\% \\
				
				& \texttt{ALS} & 13.47\% & 13.93\% & 14.13\% & 14.88\% & 14.87\% \\
				& \texttt{Fixed} & \textcolor{g}{\textbf{14.18\%}} & 13.63\% & \textcolor{g}{\textbf{15.26\%}} & \textcolor{g}{15.82\%} & \textcolor{g}{15.77\%} \\
				& \texttt{Uni} & 13.74\% & 13.26\% & 13.61\% & 14.18\% & 15.07\% \\	
				& \texttt{RandUni} & 13.73\% & 13.85\% & 14.23\% & 13.99\% & \textcolor{g}{15.69\%} \\	
				& \texttt{InvUnc} & 13.29\% & 13.58\% & 13.82\% & 14.03\% & \textcolor{g}{15.69\%} \\	
				& \texttt{cDDM} & 13.82\% & \textcolor{g}{\textbf{14.74\%}} & \textcolor{g}{15.12\%} & 15.63\% & 15.47\% \\	
				& \texttt{cEDDM} & 13.17\% & 13.41\% & 14.25\% & 14.20\% & 15.38\% \\
				& \texttt{WinErr} & 13.71\% & \textcolor{g}{14.20\%} & \textcolor{g}{14.99\%} & \textcolor{g}{\textbf{15.85\%}} & \textcolor{g}{\textbf{15.84\%}} \\
				\bottomrule
		\end{tabular}}
		
		\hspace{1em}   
		
		\subfloat{\begin{tabular}[H]{lccccccc}	
				\midrule
				Stream & Strategy & $B=1\%$ & $B=5\%$ & $B=10\%$ & $B=20\%$ & $B=50\%$ \\
				\midrule
				\multirow{9}{*}{Usenet} 
				& \texttt{ALRV} & 50.92\% & 51.36\% & 51.05\% & \textbf{52.29\%} & \textbf{54.26\%} \\ 
				& \texttt{ALR} & 50.78\% & 50.73\% & 51.44\% & 52.10\% & 53.92\% \\
				
				& \texttt{ALS} & 50.82\% & 51.31\% & 51.57\% & 51.88\% & 53.27\% \\
				& \texttt{Fixed} & \textcolor{g}{\textbf{51.94\%}} & \textcolor{g}{51.85\%} & \textcolor{g}{51.69\%} & 51.72\% & 53.66\% \\
				& \texttt{Uni} & \textcolor{g}{51.65\%} & \textcolor{g}{51.40\%} & 51.41\% & 51.80\% & 53.62\% \\	
				& \texttt{RandUni} & \textcolor{g}{51.87\%} & \textcolor{g}{51.65\%} & \textcolor{g}{51.98\%} & 52.02\% & 53.07\% \\	
				& \texttt{InvUnc} & \textcolor{g}{50.96\%} & \textcolor{g}{51.54\%} & 51.25\% & 51.69\% & 53.53\% \\	
				& \texttt{cDDM} & \textcolor{g}{51.63\%} & 50.96\% & \textcolor{g}{51.72\%} & 52.25\% & 53.24\% \\	
				& \texttt{cEDDM} & \textcolor{g}{51.29\%} & \textcolor{g}{\textbf{52.07\%}} & 51.29\% & 51.65\% & 52.47\% \\
				& \texttt{WinErr} & \textcolor{g}{51.30\%} & \textcolor{g}{51.91\%} & \textcolor{g}{\textbf{52.34\%}} & 51.27\% & 51.61\% \\
				\midrule
				\multirow{9}{*}{Gas} 
				& \texttt{ALRV} & 52.54\% & 56.22\% & 54.76\% & 54.80\% & \textbf{57.32\%} \\ 
				& \texttt{ALR} & 52.75\% & 54.00\% & 54.36\% & 53.85\% & 56.60\% \\
				
				& \texttt{ALS} & 53.62\% & 53.69\% & 54.93\% & 52.47\% & 56.02\% \\
				& \texttt{Fixed} & \textcolor{g}{54.49\%} & 54.87\% & \textcolor{g}{\textbf{55.23\%}} & 54.77\% & 54.84\% \\
				& \texttt{Uni} & 50.08\% & 53.16\% & 51.66\% & 53.59\% & 55.15\% \\	
				& \texttt{RandUni} & 50.86\% & 51.75\% & 50.19\% & 49.12\% & 53.32\% \\	
				& \texttt{InvUnc} & 46.44\% & 49.71\% & 51.41\% & 47.93\% & 52.85\% \\	
				& \texttt{cDDM} & 46.90\% & 49.86\% & 52.61\% & 48.38\% & 53.98\% \\	
				& \texttt{cEDDM} & \textcolor{g}{\textbf{54.80\%}} & \textcolor{g}{\textbf{57.88\%}} & \textcolor{g}{55.20\%} & \textcolor{g}{\textbf{55.74\%}} & 53.38\% \\
				& \texttt{WinErr} & \textcolor{g}{53.85\%} & 54.38\% & \textcolor{g}{55.03\%} & 54.33\% & 54.91\% \\
				\midrule
				\multirow{9}{*}{Poker} 
				& \texttt{ALRV} & \textbf{59.41\%} & 59.71\% & 59.79\% & 62.04\% & 65.02\% \\ 
				& \texttt{ALR} & 52.23\% & 58.72\% & 60.81\% & 61.89\% & 65.01\% \\
				
				& \texttt{ALS} & 61.05\% & 60.26\% & 60.64\% & 61.32\% & 64.91\% \\
				& \texttt{Fixed} & 53.64\% & \textcolor{g}{\textbf{63.48\%}} & \textcolor{g}{\textbf{65.10\%}} & \textcolor{g}{\textbf{69.56\%}} & \textcolor{g}{\textbf{73.32\%}} \\
				& \texttt{Uni} & 56.46\% & 49.15\% & 50.19\% & 52.10\% & 57.74\% \\	
				& \texttt{RandUni} & 53.42\% & 46.49\% & 46.25\% & 50.09\% & 58.14\% \\	
				& \texttt{InvUnc} & 51.49\% & 52.31\% & 48.24\% & 47.93\% & 54.51\% \\	
				& \texttt{cDDM} & 53.33\% & 52.26\% & 51.22\% & 52.40\% & 55.94\% \\	
				& \texttt{cEDDM} & 54.20\% & 50.13\% & 49.88\% & 50.50\% & 59.34\% \\
				& \texttt{WinErr} & 42.26\% & 47.14\% & 56.61\% & \textcolor{g}{62.18\%} & \textcolor{g}{70.83\%} \\
				\midrule
				\multirow{9}{*}{Elec} 
				& \texttt{ALRV} & \textbf{75.08\%} & \textbf{75.32\%} & \textbf{76.86\%} & \textbf{78.39\%} & 81.24\% \\ 
				& \texttt{ALR} & 72.12\% & 75.86\% & 73.80\% & 78.50\% & 81.01\% \\
				
				& \texttt{ALS} & 75.19\% & 75.14\% & 76.93\% & 78.06\% & 81.36\% \\
				& \texttt{Fixed} & 63.45\% & 73.43\% & 75.53\% & 77.62\% & \textcolor{g}{81.40\%} \\
				& \texttt{Uni} & 65.78\% & 71.49\% & 73.98\% & 74.59\% & 79.86\% \\	
				& \texttt{RandUni} & 65.63\% & 64.81\% & 68.53\% & 70.63\% & 79.13\% \\	
				& \texttt{InvUnc} & 66.65\% & 69.22\% & 69.04\% & 73.86\% & 78.44\% \\	
				& \texttt{cDDM} & 67.18\% & 71.67\% & 73.25\% & 74.72\% & 79.67\% \\	
				& \texttt{cEDDM} & 73.28\% & 74.63\% & 75.72\% & 76.93\% & 79.75\% \\
				& \texttt{WinErr} & 66.20\% & 70.82\% & 73.58\% & 78.27\% & \textcolor{g}{\textbf{81.56\%}} \\
				\bottomrule
	\end{tabular}}}
	\label{tab:aht-real}
\end{table*}

\begin{table*}[]
	\caption{RHT -- the average accuracy for real streams given a budget.}
	\centering
	
	\scalebox{0.54}{
		\subfloat{\begin{tabular}[H]{lccccccc}	
				\midrule
				Stream & Strategy & $B=1\%$ & $B=5\%$ & $B=10\%$ & $B=20\%$ & $B=50\%$ \\
				\midrule
				\multirow{9}{*}{Sensor} 
				& \texttt{ALRV} & 12.93\% & 24.24\% & 29.72\% & \textbf{39.22\%} & 46.27\% \\
				& \texttt{ALR} & 12.93\% & 23.50\% & 29.27\% & 37.56\% & 47.36\% \\
				
				& \texttt{ALS} & 12.48\% & 24.29\% & \textbf{31.07\%} & 37.02\% & \textbf{48.06\%} \\
				& \texttt{Fixed} & \textcolor{g}{16.83\%} & \textcolor{g}{25.26\%} & 27.31\% & 32.27\% & 45.09\% \\
				& \texttt{Uni} & \textcolor{g}{17.59\%} & 24.16\% & 28.48\% & 33.56\% & 46.22\% \\
				& \texttt{RandUni} & \textcolor{g}{16.93\%} & 23.67\% & 27.50\% & 32.73\% & 45.21\% \\
				& \texttt{InvUnc} & \textcolor{g}{17.47\%} & \textcolor{g}{25.69\%} & 29.11\% & 34.91\% & 45.00\% \\
				& \texttt{cDDM} & \textcolor{g}{16.83\%} & \textcolor{g}{25.11\%} & 26.57\% & 33.05\% & 46.61\% \\
				& \texttt{cEDDM} & \textcolor{g}{17.51\%} & \textcolor{g}{\textbf{26.34\%}} & 29.68\% & 33.14\% & 45.37\% \\
				& \texttt{cWin} & \textcolor{g}{\textbf{18.06\%}} & \textcolor{g}{24.46\%} & 28.08\% & 33.77\% & 46.76\% \\
				\midrule
				\multirow{9}{*}{Cover} 
				& \texttt{ALRV} & 52.56\% & 51.19\% & 51.57\% & 58.23\% & 64.01\% \\
				& \texttt{ALR} & 48.85\% & 50.80\% & 53.35\% & 57.57\% & 62.79\% \\
				
				& \texttt{ALS} & 50.79\% & 50.38\% & 52.25\% & 58.64\% & 64.08\% \\
				& \texttt{Fixed} & 51.54\% & \textcolor{g}{52.84\%} & 52.29\% & \textcolor{g}{60.73\%} & 62.18\% \\
				& \texttt{Uni} & 42.49\% & 45.12\% & 45.01\% & 55.47\% & 61.08\% \\
				& \texttt{RandUni} & 50.87\% & 50.87\% & \textcolor{g}{53.78\%} & 44.54\% & 59.92\% \\
				& \texttt{InvUnc} & \textcolor{g}{59.91\%} & \textcolor{g}{65.59\%} & \textcolor{g}{63.82\%} & \textcolor{g}{69.05\%} & \textcolor{g}{68.87\%} \\
				& \texttt{cDDM} & \textcolor{g}{\textbf{59.93\%}} & \textcolor{g}{66.01\%} & \textcolor{g}{61.30\%} & \textcolor{g}{67.59\%} & \textcolor{g}{70.09\%} \\
				& \texttt{cEDDM} & 42.32\% & \textcolor{g}{\textbf{66.97\%}} & \textcolor{g}{\textbf{67.79\%}} & \textcolor{g}{68.93\%} & \textcolor{g}{70.07\%} \\
				& \texttt{cWin} & \textcolor{g}{58.05\%} & \textcolor{g}{65.51\%} & \textcolor{g}{67.23\%} & \textcolor{g}{\textbf{70.91\%}} & \textcolor{g}{\textbf{70.82\%}} \\
				\midrule
				\multirow{9}{*}{Spam} 
				& \texttt{ALRV} & 30.74\% & 38.29\% & 53.61\% & 67.42\% & 78.54\% \\
				& \texttt{ALR} & 30.63\% & 38.33\% & 50.64\% & 67.57\% & 79.19\% \\
				
				& \texttt{ALS} & 30.76\% & 38.36\% & 54.01\% & 67.96\% & 79.09\% \\
				& \texttt{Fixed} & 26.80\% & \textcolor{g}{51.39\%} & \textcolor{g}{62.05\%} & 52.32\% & 76.79\% \\
				& \texttt{Uni} & 27.51\% & \textcolor{g}{49.30\%} & \textcolor{g}{61.84\%} & 64.09\% & 74.96\% \\
				& \texttt{RandUni} & 27.13\% & 27.93\% & \textcolor{g}{62.31\%} & 37.12\% & 73.06\% \\
				& \texttt{InvUnc} & \textcolor{g}{53.26\%} & \textcolor{g}{69.44\%} & \textcolor{g}{\textbf{82.43\%}} & \textcolor{g}{85.12\%} & \textcolor{g}{87.07\%} \\
				& \texttt{cDDM} & \textcolor{g}{\textbf{55.22\%}} & \textcolor{g}{73.68\%} & \textcolor{g}{77.25\%} & \textcolor{g}{\textbf{85.80\%}} & \textcolor{g}{\textbf{90.30\%}} \\
				& \texttt{cEDDM} & \textcolor{g}{51.82\%} & \textcolor{g}{72.47\%} & \textcolor{g}{77.79\%} & \textcolor{g}{84.18\%} & \textcolor{g}{86.93\%} \\
				& \texttt{cWin} & \textcolor{g}{52.53\%} & \textcolor{g}{\textbf{74.16\%}} & \textcolor{g}{75.23\%} & \textcolor{g}{85.41\%} & \textcolor{g}{90.03\%} \\
				\midrule
				\multirow{9}{*}{Power} 
				& \texttt{ALRV} & 4.13\% & 3.80\% & 3.96\% & 4.05\% & 3.90\% \\
				& \texttt{ALR} & 4.14\% & 4.14\% & 4.05\% & 4.03\% & 3.96\% \\
				
				& \texttt{ALS} & 4.16\% & 4.02\% & 3.98\% & 3.06\% & 3.56\% \\
				& \texttt{Fixed} & 4.09\% & 3.82\% & 3.99\% & 4.02\% & \textcolor{g}{4.03\%} \\
				& \texttt{Uni} & \textcolor{g}{4.17\%} & 4.14\% & \textcolor{g}{4.16\%} & \textcolor{g}{4.16\%} & \textcolor{g}{4.12\%} \\
				& \texttt{RandUni} & \textcolor{g}{4.17\%} & \textcolor{g}{4.16\%} & \textcolor{g}{4.17\%} & \textcolor{g}{4.17\%} & \textcolor{g}{4.17\%} \\
				& \texttt{InvUnc} & \textcolor{g}{12.58\%} & \textcolor{g}{13.87\%} & \textcolor{g}{13.53\%} & \textcolor{g}{14.75\%} & \textcolor{g}{\textbf{15.61\%}} \\
				& \texttt{cDDM} & \textcolor{g}{13.57\%} & \textcolor{g}{\textbf{14.36\%}} & \textcolor{g}{\textbf{15.27\%}} & \textcolor{g}{15.19\%} & \textcolor{g}{15.38\%} \\
				& \texttt{cEDDM} & \textcolor{g}{13.40\%} & \textcolor{g}{13.97\%} & \textcolor{g}{14.39\%} & \textcolor{g}{14.13\%} & \textcolor{g}{15.41\%} \\
				& \texttt{cWin} & \textcolor{g}{\textbf{13.98\%}} & \textcolor{g}{14.01\%} & \textcolor{g}{14.82\%} & \textcolor{g}{\textbf{15.66\%}} & \textcolor{g}{15.43\%} \\
				\bottomrule
		\end{tabular}}
		
		\hspace{1em}   
		
		\subfloat{\begin{tabular}[H]{lccccccc}	
				\midrule
				Stream & Strategy & $B=1\%$ & $B=5\%$ & $B=10\%$ & $B=20\%$ & $B=50\%$ \\
				\midrule
				\multirow{9}{*}{Usenet} 
				& \texttt{ALRV} & 51.19\% & 51.30\% & 51.12\% & 50.74\% & 52.09\% \\
				& \texttt{ALR} & 51.04\% & 51.15\% & 51.06\% & 50.11\% & 50.62\% \\
				
				& \texttt{ALS} & 51.13\% & 51.42\% & 51.24\% & 50.82\% & 50.71\% \\
				& \texttt{Fixed} & 51.18\% & 51.12\% & 51.09\% & \textcolor{g}{51.27\%} & 50.92\% \\
				& \texttt{Uni} & 51.12\% & \textcolor{g}{\textbf{51.60\%}} & 51.18\% & \textcolor{g}{51.14\%} & 51.67\% \\
				& \texttt{RandUni} & \textcolor{g}{\textbf{51.52\%}} & 51.29\% & \textcolor{g}{51.36\%} & \textcolor{g}{51.30\%} & 51.69\% \\
				& \texttt{InvUnc} & 50.32\% & 50.65\% & 51.07\% & \textcolor{g}{51.60\%} & \textcolor{g}{\textbf{53.38\%}} \\
				& \texttt{cDDM} & 50.37\% & 50.56\% & 51.03\% & \textcolor{g}{51.32\%} & \textcolor{g}{52.74\%} \\
				& \texttt{cEDDM} & 50.36\% & 50.79\% & \textcolor{g}{\textbf{51.67\%}} & \textcolor{g}{51.38\%} & \textcolor{g}{53.35\%} \\
				& \texttt{cWin} & 50.67\% & 50.81\% & \textcolor{g}{51.43\%} & \textcolor{g}{\textbf{52.05\%}} & \textcolor{g}{52.85\%} \\
				\midrule
				\multirow{9}{*}{Gas} 
				& \texttt{ALRV} & 51.47\% & 59.91\% & 62.01\% & 61.20\% & 63.24\% \\
				& \texttt{ALR} & \textbf{51.85\%} & 58.87\% & 60.96\% & 59.24\% & 62.80\% \\
				
				& \texttt{ALS} & 51.55\% & 59.54\% & 62.47\% & 70.50\% & 61.80\% \\
				& \texttt{Fixed} & 43.28\% & 49.25\% & 47.50\% & 55.09\% & 61.02\% \\
				& \texttt{Uni} & 43.11\% & 47.63\% & 58.05\% & 56.34\% & 58.56\% \\
				& \texttt{RandUni} & 41.58\% & 48.92\% & 59.88\% & 51.71\% & 60.47\% \\
				& \texttt{InvUnc} & 49.41\% & \textcolor{g}{\textbf{66.77\%}} & \textcolor{g}{64.08\%} & \textcolor{g}{\textbf{74.82\%}} & \textcolor{g}{74.11\%} \\
				& \texttt{cDDM} & 51.30\% & 57.61\% & \textcolor{g}{65.49\%} & \textcolor{g}{73.75\%} & \textcolor{g}{75.76\%} \\
				& \texttt{cEDDM} & 46.11\% & \textcolor{g}{63.06\%} & \textcolor{g}{\textbf{68.46\%}} & \textcolor{g}{73.17\%} & \textcolor{g}{\textbf{77.78\%}} \\
				& \texttt{cWin} & 49.69\% & \textcolor{g}{62.17\%} & \textcolor{g}{67.79\%} & \textcolor{g}{71.46\%} & \textcolor{g}{75.79\%} \\
				\midrule
				\multirow{9}{*}{Poker} 
				& \texttt{ALRV} & 54.08\% & 56.48\% & \textbf{58.13\%} & \textbf{58.68\%} & \textbf{59.65\%} \\
				& \texttt{ALR} & 53.52\% & 55.73\% & 57.52\% & 58.08\% & 59.53\% \\
				
				& \texttt{ALS} & \textbf{54.43\%} & \textbf{56.60\%} & 57.80\% & 57.35\% & 58.98\% \\
				& \texttt{Fixed} & 53.01\% & 53.98\% & 54.22\% & 56.10\% & 58.63\% \\
				& \texttt{Uni} & 48.16\% & 51.65\% & 50.87\% & 54.32\% & 59.33\% \\
				& \texttt{RandUni} & 53.48\% & 50.87\% & 45.56\% & 52.85\% & 58.36\% \\
				& \texttt{InvUnc} & 47.70\% & 46.36\% & 59.45\% & 42.59\% & 49.31\% \\
				& \texttt{cDDM} & 43.94\% & 51.45\% & 49.99\% & 56.44\% & 52.37\% \\
				& \texttt{cEDDM} & 48.44\% & 49.52\% & 52.17\% & 47.69\% & 53.08\% \\
				& \texttt{cWin} & 42.29\% & 49.28\% & 51.99\% & 55.75\% & 54.74\% \\
				\midrule
				\multirow{9}{*}{Elec} 
				& \texttt{ALRV} & 62.39\% & 66.18\% & 66.87\% & 69.88\% & \textbf{73.14\%} \\
				& \texttt{ALR} & 62.87\% & 66.08\% & 68.32\% & 67.04\% & 71.36\% \\
				
				& \texttt{ALS} & 62.95\% & 64.91\% & 68.22\% & 68.52\% & 69.25\% \\
				& \texttt{Fixed} & 62.06\% & 64.12\% & 65.39\% & \textcolor{g}{70.54\%} & 71.86\% \\
				& \texttt{Uni} & 58.22\% & 61.72\% & 65.63\% & 64.73\% & 69.24\% \\
				& \texttt{RandUni} & 60.20\% & 60.29\% & 61.66\% & 66.15\% & 69.30\% \\
				& \texttt{InvUnc} & 62.30\% & 63.72\% & \textcolor{g}{\textbf{67.98\%}} & \textcolor{g}{72.07\%} & 70.25\% \\
				& \texttt{cDDM} & 61.84\% & \textcolor{g}{\textbf{66.74\%}} & \textcolor{g}{67.61\%} & \textcolor{g}{70.45\%} & 69.67\% \\
				& \texttt{cEDDM} & \textcolor{g}{\textbf{66.79\%}} & 61.44\% & 65.71\% & 68.73\% & 68.54\% \\
				& \texttt{cWin} & \textcolor{g}{63.41\%} & 64.04\% & 64.42\% & \textcolor{g}{\textbf{72.42\%}} & 70.87\% \\
				\bottomrule
		\end{tabular}}
	}
	\label{tab:rht-real}
\end{table*}

\begin{table*}[]
	\caption{RCD -- the average accuracy for real streams given a budget.}
	\centering
	
	\scalebox{0.54}{
		\subfloat{\begin{tabular}[H]{lccccccc}	
				\midrule
				Stream & Strategy & $B=1\%$ & $B=5\%$ & $B=10\%$ & $B=20\%$ & $B=50\%$ \\
				\midrule
				\multirow{9}{*}{Sensor} 
				& \texttt{ALRV} & 7.20\% & \textbf{8.53\%} & \textbf{8.78\%} & \textbf{9.01\%} & \textbf{9.21\%} \\
				& \texttt{ALR} & 6.92\% & 7.31\% & 7.42\% & 7.72\% & 7.80\% \\
				
				& \texttt{ALS} & \textbf{7.26\%} & 8.19\% & 8.77\% & 8.51\% & 8.32\% \\
				& \texttt{Fixed} & 3.09\% & 3.60\% & 4.62\% & 6.19\% & 7.15\% \\
				& \texttt{Uni} & 2.40\% & 3.48\% & 3.23\% & 4.79\% & 5.15\% \\
				& \texttt{RandUni} & 2.44\% & 3.33\% & 3.77\% & 3.61\% & 5.09\% \\
				& \texttt{InvUnc} & 2.50\% & 3.24\% & 4.44\% & 4.82\% & 6.16\% \\
				& \texttt{cDDM} & 2.66\% & 4.23\% & 4.68\% & 5.83\% & 7.31\% \\
				& \texttt{cEDDM} & 3.50\% & 3.61\% & 3.70\% & 3.92\% & 5.19\% \\
				& \texttt{cWin} & 2.83\% & 4.32\% & 4.76\% & 6.04\% & 7.00\% \\
				\midrule
				\multirow{9}{*}{Cover} 
				& \texttt{ALRV} & \textbf{58.01\%} & 60.11\% & 60.57\% & 60.72\% & 61.67\% \\
				& \texttt{ALR} & 57.57\% & 59.87\% & 60.10\% & 60.32\% & 61.24\% \\
				
				& \texttt{ALS} & 57.63\% & 60.08\% & 60.42\% & 60.41\% & 62.22\% \\
				& \texttt{Fixed} & 32.17\% & \textcolor{g}{\textbf{65.10\%}} & \textcolor{g}{62.11\%} & \textcolor{g}{\textbf{65.14\%}} & \textcolor{g}{74.75\%} \\
				& \texttt{Uni} & 31.36\% & 52.47\% & \textcolor{g}{\textbf{62.33\%}} & \textcolor{g}{63.64\%} & \textcolor{g}{73.40\%} \\
				& \texttt{RandUni} & 36.76\% & 53.81\% & 54.48\% & 37.06\% & \textcolor{g}{72.39\%} \\
				& \texttt{InvUnc} & 6.93\% & 33.01\% & 58.79\% & \textcolor{g}{63.16\%} & \textcolor{g}{69.96\%} \\
				& \texttt{cDDM} & 31.02\% & 57.74\% & 60.08\% & \textcolor{g}{61.66\%} & \textcolor{g}{71.34\%} \\
				& \texttt{cEDDM} & 52.66\% & 37.25\% & 55.18\% & \textcolor{g}{65.12\%} & \textcolor{g}{\textbf{77.22\%}} \\
				& \texttt{cWin} & 31.82\% & 60.06\% & 58.84\% & \textcolor{g}{63.36\%} & \textcolor{g}{73.26\%} \\
				\midrule
				\multirow{9}{*}{Spam} 
				& \texttt{ALRV} & 27.55\% & 55.57\% & 66.74\% & 81.61\% & 88.53\% \\
				& \texttt{ALR} & 27.53\% & 52.49\% & 65.65\% & \textbf{81.80\%} & 88.28\% \\
				
				& \texttt{ALS} & 27.54\% & 54.76\% & \textbf{70.31\%} & 80.08\% & 86.94\% \\
				& \texttt{Fixed} & \textcolor{g}{33.45\%} & 36.70\% & 41.95\% & 73.55\% & 84.91\% \\
				& \texttt{Uni} & 25.28\% & 25.84\% & 59.50\% & 64.50\% & 87.56\% \\
				& \texttt{RandUni} & 25.59\% & \textcolor{g}{\textbf{57.00\%}} & 53.30\% & 62.96\% & 87.79\% \\
				& \texttt{InvUnc} & 24.08\% & 34.94\% & 46.58\% & 50.31\% & 84.00\% \\
				& \texttt{cDDM} & \textcolor{g}{40.47\%} & 41.37\% & 43.97\% & 73.50\% & 82.50\% \\
				& \texttt{cEDDM} & \textcolor{g}{27.63\%} & 55.03\% & 36.15\% & 80.54\% & \textcolor{g}{\textbf{88.62\%}} \\
				& \texttt{cWin} & \textcolor{g}{\textbf{40.59\%}} & 41.27\% & 42.45\% & 71.90\% & 84.08\% \\
				\midrule
				\multirow{9}{*}{Power} 
				& \texttt{ALRV} & 13.48\% & 13.68\% & 15.10\% & 15.23\% & 15.74\% \\
				& \texttt{ALR} & 13.44\% & 13.81\% & \textbf{15.40\%} & 15.50\% & 15.49\% \\
				
				& \texttt{ALS} & 13.54\% & 13.70\% & 15.07\% & 15.27\% & 13.71\% \\
				& \texttt{Fixed} & \textcolor{g}{13.58\%} & 13.78\% & 14.81\% & \textcolor{g}{15.63\%} & 15.61\% \\
				& \texttt{Uni} & \textcolor{g}{\textbf{13.97\%}} & \textcolor{g}{13.97\%} & 14.18\% & 14.47\% & 14.94\% \\
				& \texttt{RandUni} & \textcolor{g}{13.68\%} & 13.78\% & 13.73\% & 14.71\% & 14.69\% \\
				& \texttt{InvUnc} & \textcolor{g}{13.67\%} & \textcolor{g}{13.83\%} & 13.56\% & 14.48\% & \textcolor{g}{\textbf{15.84\%}} \\
				& \texttt{cDDM} & \textcolor{g}{13.60\%} & \textcolor{g}{14.48\%} & 14.67\% & \textcolor{g}{15.53\%} & \textcolor{g}{15.82\%} \\
				& \texttt{cEDDM} & 13.33\% & \textcolor{g}{13.98\%} & 14.54\% & 14.62\% & 15.09\% \\
				& \texttt{cWin} & \textcolor{g}{13.70\%} & \textcolor{g}{\textbf{14.54\%}} & 14.92\% & \textcolor{g}{\textbf{15.66\%}} & \textcolor{g}{15.76\%} \\
				\bottomrule
		\end{tabular}}
		
		\hspace{1em}   
		
		\subfloat{\begin{tabular}[H]{lccccccc}	
				\midrule
				Stream & Strategy & $B=1\%$ & $B=5\%$ & $B=10\%$ & $B=20\%$ & $B=50\%$ \\
				\midrule
				\multirow{9}{*}{Usenet} 
				& \texttt{ALRV} & 50.79\% & 50.74\% & 49.92\% & 52.36\% & 54.46\% \\
				& \texttt{ALR} & 50.73\% & 50.66\% & 51.51\% & 50.60\% & \textbf{55.00\%} \\
				
				& \texttt{ALS} & 50.69\% & 50.97\% & 51.54\% & 52.28\% & 53.12\% \\
				& \texttt{Fixed} & \textcolor{g}{\textbf{51.80\%}} & \textcolor{g}{\textbf{51.76\%}} & \textcolor{g}{51.69\%} & 52.14\% & 52.73\% \\
				& \texttt{Uni} & \textcolor{g}{51.30\%} & \textcolor{g}{51.23\%} & \textcolor{g}{51.56\%} & 52.20\% & 53.09\% \\
				& \texttt{RandUni} & \textcolor{g}{51.38\%} & \textcolor{g}{51.61\%} & 51.45\% & 51.96\% & 52.65\% \\
				& \texttt{InvUnc} & \textcolor{g}{\textbf{51.80\%}} & \textcolor{g}{51.69\%} & \textcolor{g}{52.05\%} & \textcolor{g}{\textbf{52.54\%}} & 53.27\% \\
				& \texttt{cDDM} & \textcolor{g}{51.10\%} & \textcolor{g}{51.14\%} & \textcolor{g}{\textbf{52.34\%}} & 52.11\% & 52.07\% \\
				& \texttt{cEDDM} & 50.48\% & \textcolor{g}{51.16\%} & 51.49\% & 51.87\% & 52.93\% \\
				& \texttt{cWin} & \textcolor{g}{51.29\%} & \textcolor{g}{51.45\%} & \textcolor{g}{51.76\%} & \textcolor{g}{52.38\%} & 52.29\% \\
				\midrule
				\multirow{9}{*}{Gas} 
				& \texttt{ALRV} & 48.48\% & 57.15\% & 57.05\% & 59.07\% & \textbf{63.20\%} \\
				& \texttt{ALR} & 49.57\% & \textbf{58.04\%} & 56.69\% & 60.07\% & 63.08\% \\
				
				& \texttt{ALS} & 49.24\% & 53.60\% & \textbf{58.50\%} & \textbf{61.64\%} & 63.14\% \\
				& \texttt{Fixed} & \textcolor{g}{\textbf{55.00\%}} & 51.68\% & 52.74\% & 52.64\% & 53.21\% \\
				& \texttt{Uni} & \textcolor{g}{50.72\%} & 50.32\% & 51.73\% & 53.28\% & 52.13\% \\
				& \texttt{RandUni} & \textcolor{g}{49.63\%} & 49.14\% & 52.87\% & 53.40\% & 50.60\% \\
				& \texttt{InvUnc} & \textcolor{g}{49.78\%} & 51.41\% & 50.86\% & 52.72\% & 51.65\% \\
				& \texttt{cDDM} & \textcolor{g}{51.09\%} & 51.12\% & 51.05\% & 52.62\% & 53.82\% \\
				& \texttt{cEDDM} & \textcolor{g}{52.06\%} & 48.52\% & 50.99\% & 48.91\% & 51.03\% \\
				& \texttt{cWin} & \textcolor{g}{50.09\%} & 50.89\% & 51.02\% & 52.91\% & 53.73\% \\
				\midrule
				\multirow{9}{*}{Poker} 
				& \texttt{ALRV} & \textbf{54.76\%} & 56.90\% & 58.53\% & 58.73\% & 59.32\% \\
				& \texttt{ALR} & 54.00\% & 57.93\% & 58.98\% & 58.87\% & 59.45\% \\
				
				& \texttt{ALS} & 53.91\% & 57.96\% & 58.50\% & 58.83\% & 59.38\% \\
				& \texttt{Fixed} & 50.87\% & \textcolor{g}{\textbf{59.97\%}} & \textcolor{g}{\textbf{64.32\%}} & \textcolor{g}{\textbf{65.48\%}} & \textcolor{g}{71.57\%} \\
				& \texttt{Uni} & 41.80\% & 52.41\% & 57.18\% & 52.41\% & \textcolor{g}{\textbf{74.73\%}} \\
				& \texttt{RandUni} & 35.11\% & 41.22\% & 50.00\% & 57.22\% & \textcolor{g}{71.10\%} \\
				& \texttt{InvUnc} & 44.08\% & 44.57\% & 42.47\% & 47.37\% & \textcolor{g}{62.65\%} \\
				& \texttt{cDDM} & 47.64\% & 43.88\% & 53.93\% & \textcolor{g}{59.11\%} & \textcolor{g}{70.50\%} \\
				& \texttt{cEDDM} & 38.91\% & 47.07\% & 50.40\% & 56.57\% & \textcolor{g}{74.14\%} \\
				& \texttt{cWin} & 44.70\% & 49.64\% & \textcolor{g}{58.99\%} & \textcolor{g}{61.81\%} & \textcolor{g}{72.96\%} \\
				\midrule
				\multirow{9}{*}{Elec} 
				& \texttt{ALRV} & 61.60\% & 68.96\% & 70.84\% & 72.08\% & 73.78\% \\
				& \texttt{ALR} & 61.36\% & 67.20\% & 70.96\% & 72.14\% & 72.82\% \\
				
				& \texttt{ALS} & 63.03\% & 69.15\% & 70.84\% & 71.23\% & 72.67\% \\
				& \texttt{Fixed} & \textcolor{g}{\textbf{67.90\%}} & 68.22\% & \textcolor{g}{\textbf{75.03\%}} & \textcolor{g}{75.99\%} & \textcolor{g}{79.11\%} \\
				& \texttt{Uni} & \textcolor{g}{67.26\%} & \textcolor{g}{73.96\%} & \textcolor{g}{74.59\%} & 72.02\% & \textcolor{g}{77.94\%} \\
				& \texttt{RandUni} & \textcolor{g}{64.42\%} & 64.01\% & \textcolor{g}{74.32\%} & 70.59\% & \textcolor{g}{77.45\%} \\
				& \texttt{InvUnc} & 58.36\% & 62.39\% & \textcolor{g}{71.93\%} & \textcolor{g}{75.69\%} & \textcolor{g}{77.79\%} \\
				& \texttt{cDDM} & \textcolor{g}{65.71\%} & \textcolor{g}{71.31\%} & \textcolor{g}{73.92\%} & \textcolor{g}{75.62\%} & \textcolor{g}{78.15\%} \\
				& \texttt{cEDDM} & \textcolor{g}{67.62\%} & \textcolor{g}{\textbf{74.92\%}} & \textcolor{g}{73.81\%} & \textcolor{g}{\textbf{76.25\%}} & \textcolor{g}{\textbf{80.27\%}} \\
				& \texttt{cWin} & \textcolor{g}{65.62\%} & \textcolor{g}{70.22\%} & \textcolor{g}{72.99\%} & \textcolor{g}{74.86\%} & \textcolor{g}{77.97\%} \\
				\bottomrule
		\end{tabular}}
	}
	\label{tab:rcd-real}
\end{table*}

\begin{table*}[]
	\caption{AWE -- the average accuracy for real streams given a budget.}
	\centering
	
	\scalebox{0.54}{
		\subfloat{\begin{tabular}[H]{lccccccc}	
				\midrule
				Stream & Strategy & $B=1\%$ & $B=5\%$ & $B=10\%$ & $B=20\%$ & $B=50\%$ \\
				\midrule
				\multirow{9}{*}{Sensor} 
				& \texttt{ALRV} & 2.13\% & 2.17\% & 2.15\% & 2.16\% & 2.26\% \\
				& \texttt{ALR} & 2.20\% & 2.12\% & 2.24\% & 2.21\% & \textbf{2.34\%} \\
				
				& \texttt{ALS} & 2.03\% & 2.13\% & 2.24\% & 2.18\% & 2.17\% \\
				& \texttt{Fixed} & 2.00\% & \textcolor{g}{2.26\%} & \textcolor{g}{2.26\%} & \textcolor{g}{\textbf{2.27\%}} & 2.20\% \\
				& \texttt{Uni} & 2.14\% & \textcolor{g}{2.28\%} & 2.09\% & 2.04\% & 2.11\% \\
				& \texttt{RandUni} & 1.95\% & 1.97\% & 1.77\% & 2.09\% & 2.15\% \\
				& \texttt{InvUnc} & 2.15\% & \textcolor{g}{2.20\%} & \textcolor{g}{2.25\%} & 2.14\% & 2.16\% \\
				& \texttt{cDDM} & \textcolor{g}{\textbf{2.30\%}} & 2.10\% & 2.19\% & \textcolor{g}{2.24\%} & 2.33\% \\
				& \texttt{cEDDM} & 2.04\% & \textcolor{g}{\textbf{2.30\%}} & 2.14\% & \textcolor{g}{2.22\%} & 2.26\% \\
				& \texttt{cWin} & \textcolor{g}{2.21\%} & 2.07\% & 2.19\% & \textcolor{g}{\textbf{2.27\%}} & 2.20\% \\
				\midrule
				\multirow{9}{*}{Cover} 
				& \texttt{ALRV} & \textbf{58.88\%} & 65.04\% & 67.08\% & 66.72\% & 64.55\% \\
				& \texttt{ALR} & 55.50\% & 65.08\% & 66.94\% & 67.08\% & 64.01\% \\
				
				& \texttt{ALS} & 55.22\% & 64.41\% & 67.02\% & 66.61\% & 64.67\% \\
				& \texttt{Fixed} & 57.72\% & \textcolor{g}{\textbf{65.43\%}} & \textcolor{g}{\textbf{67.70\%}} & \textcolor{g}{\textbf{67.13\%}} & 64.52\% \\
				& \texttt{Uni} & 30.53\% & 36.57\% & 45.43\% & 60.45\% & 63.87\% \\
				& \texttt{RandUni} & 36.60\% & 36.50\% & 36.50\% & 45.14\% & 63.96\% \\
				& \texttt{InvUnc} & 39.16\% & 36.50\% & 48.90\% & 52.95\% & 63.81\% \\
				& \texttt{cDDM} & 48.10\% & 59.82\% & 63.74\% & 66.61\% & \textcolor{g}{64.97\%} \\
				& \texttt{cEDDM} & 36.31\% & 36.42\% & 43.55\% & 65.68\% & 63.56\% \\
				& \texttt{cWin} & 44.52\% & 60.03\% & 64.32\% & 65.82\% & \textcolor{g}{\textbf{65.08\%}} \\
				\midrule
				\multirow{9}{*}{Spam} 
				& \texttt{ALRV} & 50.80\% & 50.80\% & 65.99\% & 81.33\% & 88.14\% \\
				& \texttt{ALR} & 50.80\% & 50.80\% & 65.28\% & 79.10\% & 88.22\% \\
				
				& \texttt{ALS} & 50.80\% & 50.80\% & 67.55\% & 81.10\% & 88.66\% \\
				& \texttt{Fixed} & \textcolor{g}{54.18\%} & \textcolor{g}{\textbf{71.11\%}} & \textcolor{g}{73.27\%} & \textcolor{g}{86.13\%} & 81.13\% \\
				& \texttt{Uni} & 34.70\% & \textcolor{g}{52.89\%} & \textcolor{g}{80.68\%} & \textcolor{g}{83.25\%} & \textcolor{g}{90.48\%} \\
				& \texttt{RandUni} & 49.86\% & \textcolor{g}{57.04\%} & 56.96\% & \textcolor{g}{84.97\%} & 88.46\% \\
				& \texttt{InvUnc} & 30.27\% & 42.48\% & \textcolor{g}{75.48\%} & \textcolor{g}{84.05\%} & 80.33\% \\
				& \texttt{cDDM} & \textcolor{g}{53.00\%} & \textcolor{g}{69.84\%} & \textcolor{g}{\textbf{81.20\%}} & \textcolor{g}{\textbf{88.33\%}} & 81.91\% \\
				& \texttt{cEDDM} & 50.05\% & 46.81\% & 56.54\% & \textcolor{g}{85.99\%} & 87.26\% \\
				& \texttt{cWin} & \textcolor{g}{\textbf{58.27\%}} & \textcolor{g}{69.19\%} & \textcolor{g}{76.93\%} & \textcolor{g}{85.37\%} & \textcolor{g}{\textbf{90.54\%}} \\
				\midrule
				\multirow{9}{*}{Power} 
				& \texttt{ALRV} & \textbf{4.17\%} & 3.98\% & 3.78\% & 4.16\% & 4.04\% \\
				& \texttt{ALR} & 4.17\% & 4.07\% & 3.15\% & 3.53\% & 3.97\% \\
				
				& \texttt{ALS} & 4.17\% & 3.67\% & 3.90\% & 4.09\% & 3.83\% \\
				& \texttt{Fixed} & 4.17\% & 3.94\% & \textcolor{g}{3.97\%} & 3.81\% & 3.89\% \\
				& \texttt{Uni} & 4.02\% & 4.17\% & \textcolor{g}{\textbf{4.17\%}} & \textcolor{g}{4.17\%} & \textcolor{g}{4.17\%} \\
				& \texttt{RandUni} & 4.17\% & \textcolor{g}{4.08\%} & \textcolor{g}{4.03\%} & 3.99\% & \textcolor{g}{4.18\%} \\
				& \texttt{InvUnc} & 4.12\% & 3.65\% & 3.37\% & \textcolor{g}{\textbf{4.18\%}} & 4.04\% \\
				& \texttt{cDDM} & 4.17\% & \textcolor{g}{\textbf{4.17\%}} & \textcolor{g}{\textbf{4.17\%}} & 3.37\% & 3.97\% \\
				& \texttt{cEDDM} & 4.17\% & 3.77\% & 3.40\% & 3.82\% & \textcolor{g}{\textbf{4.19\%}} \\
				& \texttt{cWin} & 4.17\% & \textcolor{g}{\textbf{4.17\%}} & \textcolor{g}{\textbf{4.17\%}} & 3.59\% & 3.90\% \\
				\bottomrule
		\end{tabular}}
		
		\hspace{1em}   
		
		\subfloat{\begin{tabular}[H]{lccccccc}	
				\midrule
				Stream & Strategy & $B=1\%$ & $B=5\%$ & $B=10\%$ & $B=20\%$ & $B=50\%$ \\
				\midrule
				\multirow{9}{*}{Usenet} 
				& \texttt{ALRV} & 49.57\% & 49.57\% & 49.57\% & 49.21\% & 49.90\% \\
				& \texttt{ALR} & 49.56\% & 49.56\% & 49.56\% & 48.29\% & 49.64\% \\
				
				& \texttt{ALS} & 49.56\% & 49.56\% & 49.56\% & 48.89\% & 49.27\% \\
				& \texttt{Fixed} & 49.57\% & 49.57\% & 49.57\% & 49.06\% & \textcolor{g}{50.06\%} \\
				& \texttt{Uni} & \textcolor{g}{50.01\%} & \textcolor{g}{50.03\%} & \textcolor{g}{\textbf{50.25\%}} & \textcolor{g}{50.16\%} & \textcolor{g}{\textbf{50.78\%}} \\
				& \texttt{RandUni} & \textcolor{g}{\textbf{50.28\%}} & \textcolor{g}{\textbf{50.25\%}} & \textcolor{g}{50.23\%} & \textcolor{g}{\textbf{50.34\%}} & 50.39\% \\
				& \texttt{InvUnc} & 49.30\% & 49.57\% & 49.19\% & 49.21\% & 49.12\% \\
				& \texttt{cDDM} & 49.57\% & 49.57\% & 49.57\% & \textcolor{g}{49.35\%} & 49.63\% \\
				& \texttt{cEDDM} & 49.57\% & \textcolor{g}{50.12\%} & \textcolor{g}{50.05\%} & \textcolor{g}{49.72\%} & \textcolor{g}{50.26\%} \\
				& \texttt{cWin} & 49.57\% & 49.57\% & 49.57\% & 49.01\% & 49.79\% \\
				\midrule
				\multirow{9}{*}{Gas} 
				& \texttt{ALRV} & 13.46\% & 14.58\% & 16.74\% & 18.91\% & 17.17\% \\
				& \texttt{ALR} & 13.46\% & 16.53\% & 14.82\% & 18.79\% & 17.83\% \\
				
				& \texttt{ALS} & 13.46\% & 14.22\% & 16.59\% & 16.31\% & 18.61\% \\
				& \texttt{Fixed} & 13.46\% & 14.19\% & 16.69\% & 16.33\% & 15.37\% \\
				& \texttt{Uni} & \textcolor{g}{\textbf{16.43\%}} & \textcolor{g}{\textbf{17.48\%}} & \textcolor{g}{\textbf{19.09\%}} & 15.63\% & \textcolor{g}{22.54\%} \\
				& \texttt{RandUni} & \textcolor{g}{\textbf{16.43\%}} & \textcolor{g}{16.65\%} & 15.92\% & 15.54\% & \textcolor{g}{\textbf{20.81\%}} \\
				& \texttt{InvUnc} & \textcolor{g}{15.65\%} & 12.44\% & \textcolor{g}{18.07\%} & 17.21\% & 14.64\% \\
				& \texttt{cDDM} & 13.46\% & \textcolor{g}{17.19\%} & 13.97\% & \textcolor{g}{\textbf{20.22\%}} & 16.65\% \\
				& \texttt{cEDDM} & 13.46\% & \textcolor{g}{16.98\%} & 14.26\% & 17.28\% & \textcolor{g}{18.66\%} \\
				& \texttt{cWin} & 13.46\% & \textcolor{g}{17.18\%} & 13.85\% & 18.38\% & 15.36\% \\
				\midrule
				\multirow{9}{*}{Poker} 
				& \texttt{ALRV} & 35.29\% & 32.91\% & 34.25\% & \textbf{36.52\%} & 41.35\% \\
				& \texttt{ALR} & 36.97\% & 36.98\% & 34.37\% & 36.25\% & 40.18\% \\
				
				& \texttt{ALS} & 25.31\% & 35.72\% & 36.61\% & 35.15\% & \textbf{41.73\%} \\
				& \texttt{Fixed} & 4.77\% & 4.77\% & 15.29\% & 31.85\% & 38.88\% \\
				& \texttt{Uni} & \textcolor{g}{37.02\%} & 29.15\% & 29.34\% & 30.64\% & 37.91\% \\
				& \texttt{RandUni} & 28.05\% & 16.50\% & 29.74\% & 32.54\% & 38.12\% \\
				& \texttt{InvUnc} & 4.77\% & 4.77\% & 15.12\% & 31.97\% & 34.61\% \\
				& \texttt{cDDM} & \textcolor{g}{\textbf{38.98\%}} & \textcolor{g}{\textbf{40.68\%}} & \textcolor{g}{\textbf{39.28\%}} & 35.26\% & 37.35\% \\
				& \texttt{cEDDM} & 30.62\% & 34.96\% & 33.85\% & 29.35\% & 35.63\% \\
				& \texttt{cWin} & \textcolor{g}{37.26\%} & 9.59\% & 16.27\% & 32.82\% & 37.37\% \\
				\midrule
				\multirow{9}{*}{Elec} 
				& \texttt{ALRV} & 42.45\% & \textbf{55.89\%} & 53.11\% & 52.84\% & 50.11\% \\
				& \texttt{ALR} & 42.45\% & 55.15\% & 53.09\% & 53.52\% & 52.05\% \\
				
				& \texttt{ALS} & 42.45\% & 54.82\% & 53.54\% & 53.23\% & 54.12\% \\
				& \texttt{Fixed} & 42.45\% & 51.98\% & 45.09\% & \textcolor{g}{\textbf{57.02\%}} & \textcolor{g}{54.22\%} \\
				& \texttt{Uni} & 42.45\% & 42.45\% & 42.45\% & 48.72\% & \textcolor{g}{54.31\%} \\
				& \texttt{RandUni} & 42.45\% & 42.45\% & 42.45\% & 52.67\% & 52.04\% \\
				& \texttt{InvUnc} & 42.45\% & 42.45\% & 42.45\% & \textcolor{g}{54.40\%} & \textcolor{g}{\textbf{55.31\%}} \\
				& \texttt{cDDM} & 42.54\% & 54.11\% & \textcolor{g}{54.96\%} & \textcolor{g}{53.84\%} & 54.46\% \\
				& \texttt{cEDDM} & \textcolor{g}{\textbf{46.88\%}} & 48.90\% & \textcolor{g}{53.69\%} & 53.29\% & \textcolor{g}{54.77\%} \\
				& \texttt{cWin} & \textcolor{g}{43.78\%} & 46.65\% & \textcolor{g}{\textbf{56.02\%}} & \textcolor{g}{53.59\%} & \textcolor{g}{54.41\%} \\
				\bottomrule
		\end{tabular}}
	}
	\label{tab:awe-real}
\end{table*}

The included results for single classifiers (Tab. \ref{tab:aht-real} and \ref{tab:rht-real}) and ensembles (Tab. \ref{tab:rcd-real} and \ref{tab:awe-real}) clearly show that the proposed strategies were able to enhance the accuracy for a wide range of budgets. The single classifiers achieved similar performance on average -- for some cases, the AHT algorithm performed better, for example, Cover, Spam or Elec, but for the rest, like Sensor or Gas, the RHT classifier was more efficient. However, it can be easily noticed that the latter was improved much more frequently than the former. Although several improvements can also be seen for both RCD and AWE, there are some cases -- the Sensor stream, for instance -- in which they were not able to learn concepts properly with or without a self-labeling module, so they perform slightly worse in general than the single classifiers. It is especially interesting since the diversity of the ensembles should, theoretically, provide better adaptivity. Although it may be trivial, one should also notice that the overall accuracy rises for all classifiers when the budget increases.

\begin{figure}[t]
	\centering
	\subfloat{\includegraphics[width=0.35\columnwidth]{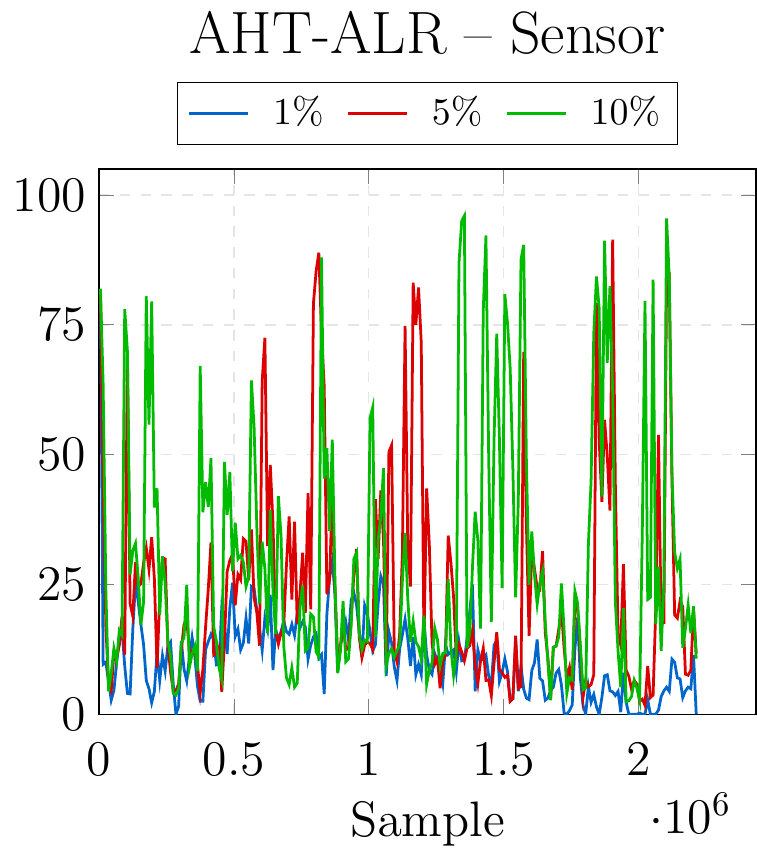}}
	\hspace{0.15cm}
	\subfloat{\includegraphics[width=0.35\columnwidth]{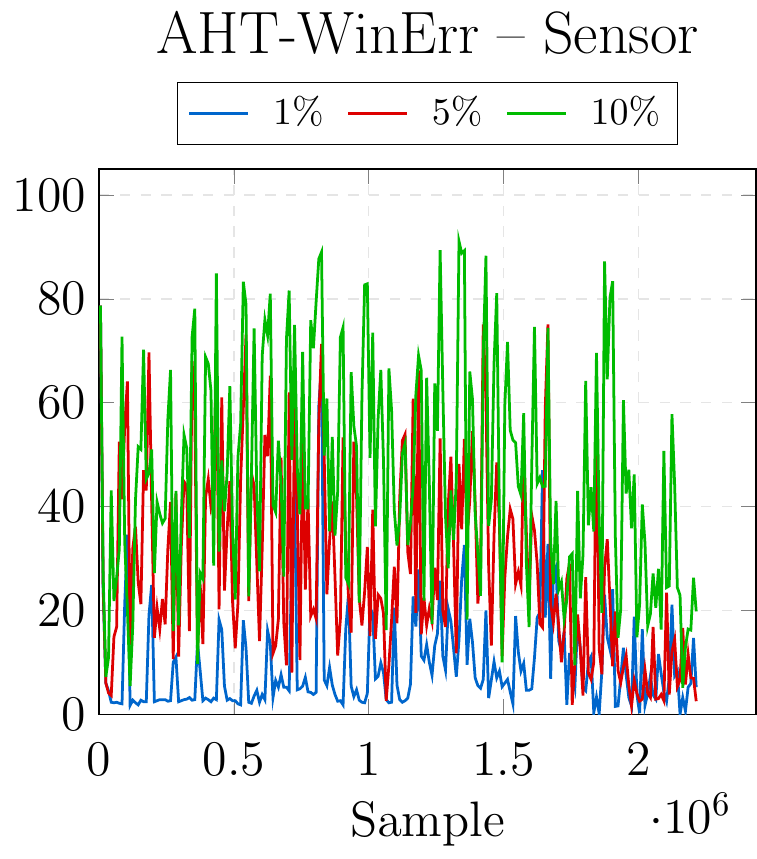}}\\
	\subfloat{\includegraphics[width=0.35\columnwidth]{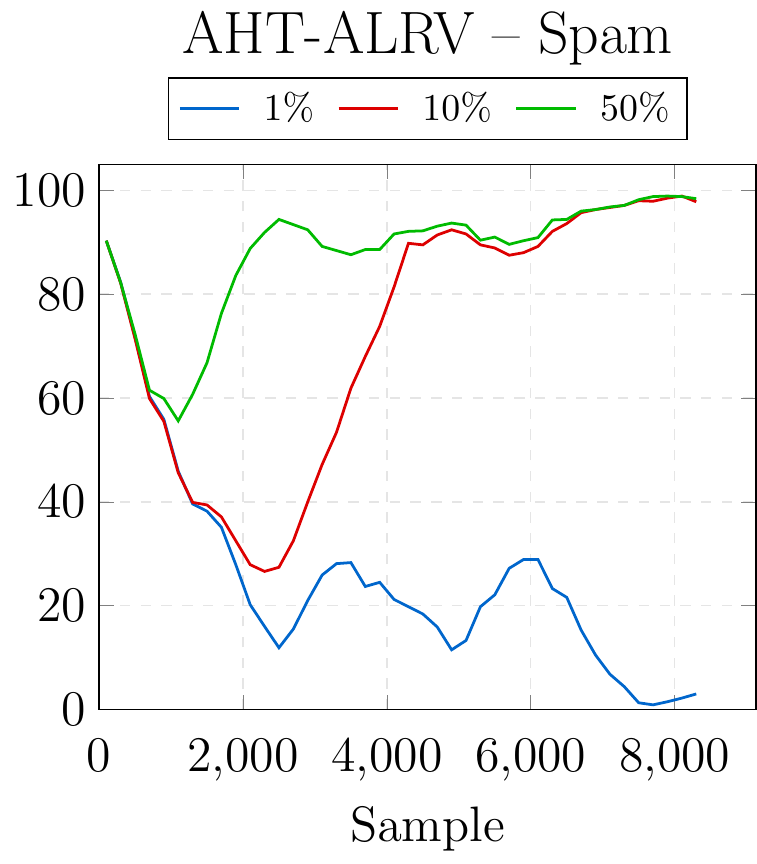}}
	\hspace{0.15cm}
	\subfloat{\includegraphics[width=0.35\columnwidth]{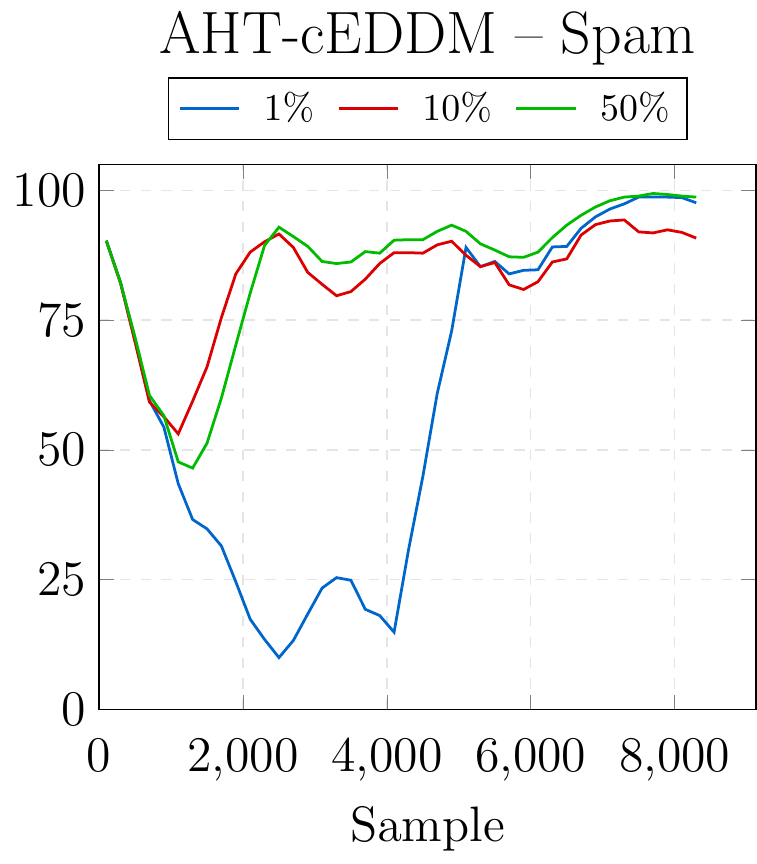}}
	\caption{Accuracy of AHT with different strategies given a budget for the Sensor and Spam data stream.}
	\label{fig:aht-real}
\end{figure}

Both single learners performed well on the Sensor stream. AHT using the \texttt{Fixed} or \texttt{WinErr} strategy improved results for almost all budgets from nearly 3\% for $B=1\%$ to more than 15\% for $B=10\%$. One should notice that there are over 50 classes for the stream, so the improvement is very significant. In Fig. \ref{fig:aht-real} it can be clearly seen that for AHT with \texttt{WinErr} and budget values equal to 5\% or 10\% the accuracy within the sliding window is more frequently on a higher level than for \texttt{ALR}. RHT boosted the learning process for almost all strategies on 1\% and 5\% budget, achieving the best results on average for such settings. Neither of the committees was able to learn the concept properly. Even for the basic approaches without self-labeling they worked practically at random, so there was no chance that a semi-supervised learning step could improve something without a sufficiently reliable model.

The most impressive improvement was obtained for the Spam stream. In the case of AHT, the enhancement is definitely present on low budgets for all strategies, excluding \texttt{Uni}. Among others, the \texttt{InvUnc} strategy gave almost 60\% gain in accuracy for $B=1\%$ and more than 30\% for $B=5\%$. \texttt{cEDDM} was able to increase the accuracy by 10\% even for $B=10\%$. The graph for the stream and the \texttt{cEDDM} strategy (Fig. \ref{fig:aht-real}) shows that \texttt{ALRV} was not able to learn new concepts quickly enough for both 1\% (after the second drift) and 10\% (after the first change). For $B=50\%$ there is no noticeable gain. RHT provides slightly lower accuracy for low budgets, however, it was able to improve the learning process for the whole range of budgets when the \texttt{InvUnc}, \texttt{cDDM}, \texttt{cEDDM} or \texttt{WinErr} strategy was used. The \texttt{InvUnc} strategy gave almost 30\% gain on $B=10\%$, \texttt{cDDM} gave nearly 20\% on $B=20\%$ and more than 12\% on $B= 50\%$, to name a few. Out of the two ensembles, AWE performed much better than RCD. The former achieved as good results as the single classifiers and even the best result (90.54\%) overall, using \texttt{WinErr} on $B=50\%$. The latter improved adaptivity mainly for the lowest budget. 

\begin{figure}[t]
	\centering
	\subfloat{\includegraphics[width=0.35\columnwidth]{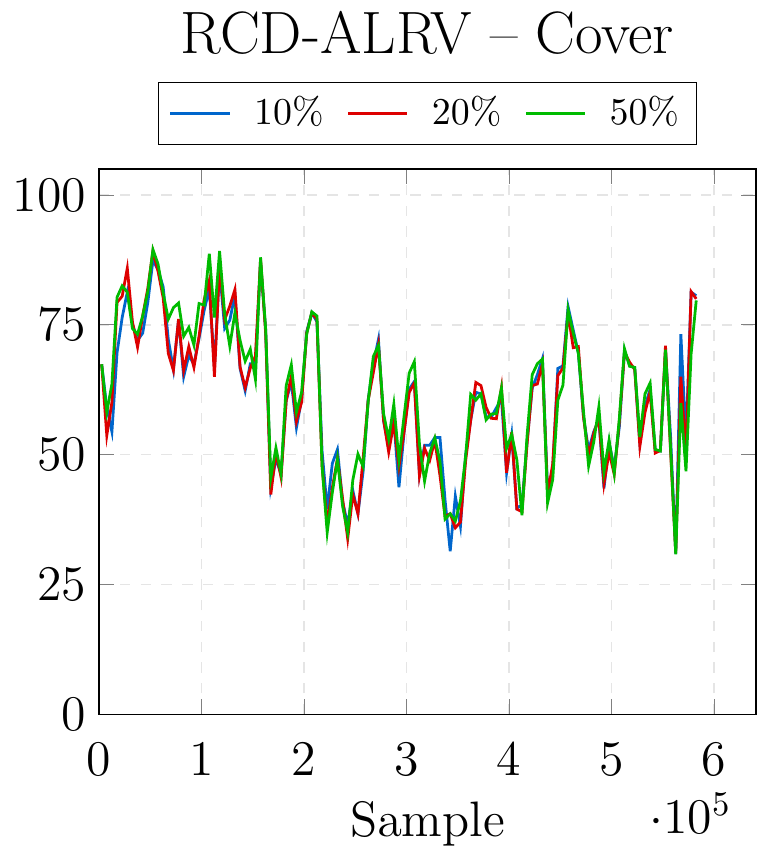}}
	\hspace{0.15cm}
	\subfloat{\includegraphics[width=0.35\columnwidth]{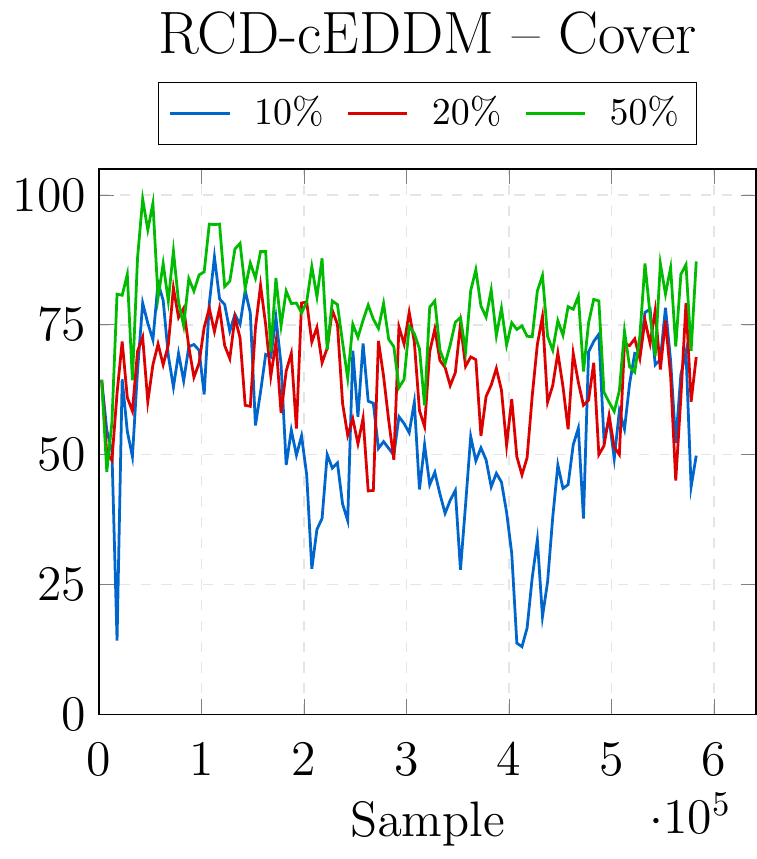}}\\
	\subfloat{\includegraphics[width=0.35\columnwidth]{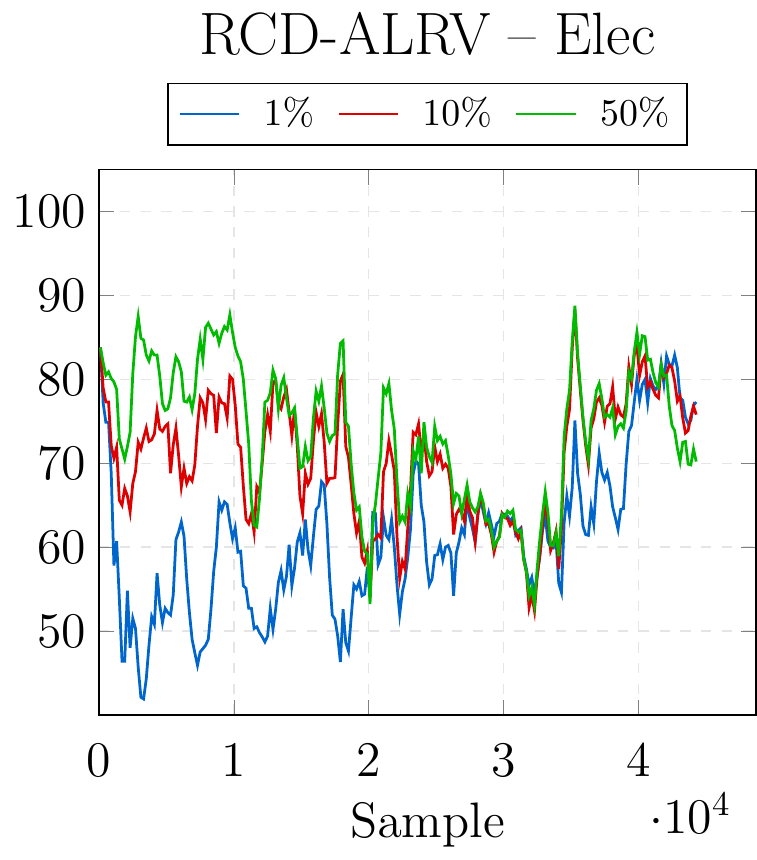}}
	\hspace{0.15cm}
	\subfloat{\includegraphics[width=0.35\columnwidth]{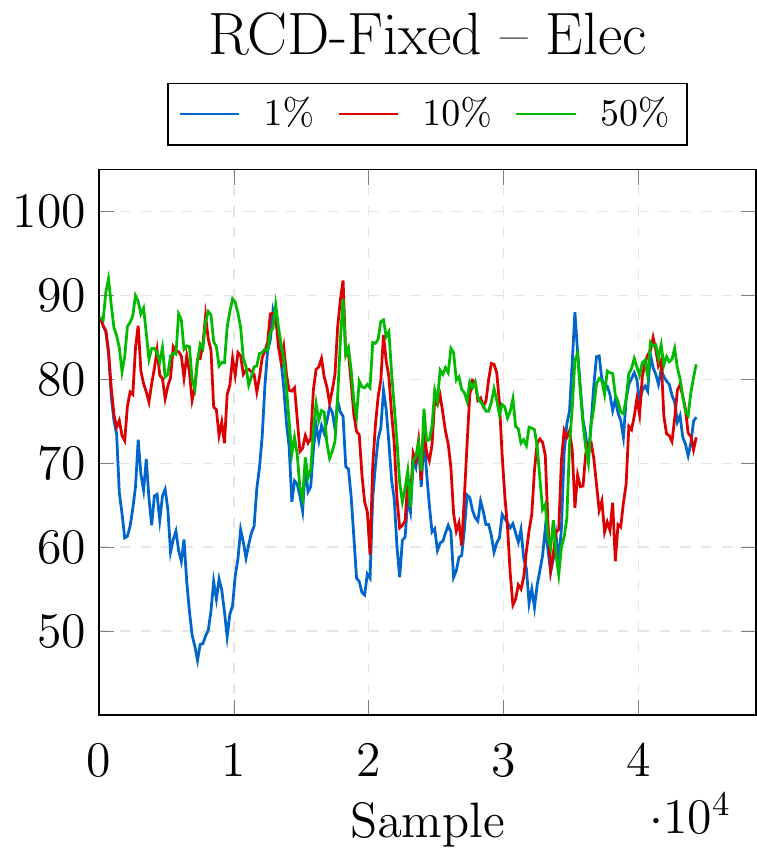}}
	\caption{Accuracy of RCD with different strategies given a budget for the Cover and Elec data stream.}
	\label{fig:rcd-real}
\end{figure}

A very similar relation between AHT and RHT, regarding an average accuracy and a rate of improvements, can be noticed for the Cover stream. AHT worked best with self-labeling approaches for $B=1\%$, while RHT for all budgets. RCD was able to significantly boost the learning mainly for high values of the budget. For $B=50\%$ almost all self-labeling strategies provided a gain from 10\% to 15\%. In Fig. \ref{fig:rcd-real} it can be seen that when $B=20\%$ the accuracy was more frequently between 50\% and 75\% than in the case of \texttt{AL} and on $B=50\%$ above $75\%$ for most of the time. The \texttt{ALRV} strategy remained unaffected by the increasing budget, even if five times more labels were provided.

\begin{figure}[ht]
	\centering
	\subfloat{\includegraphics[width=0.35\columnwidth]{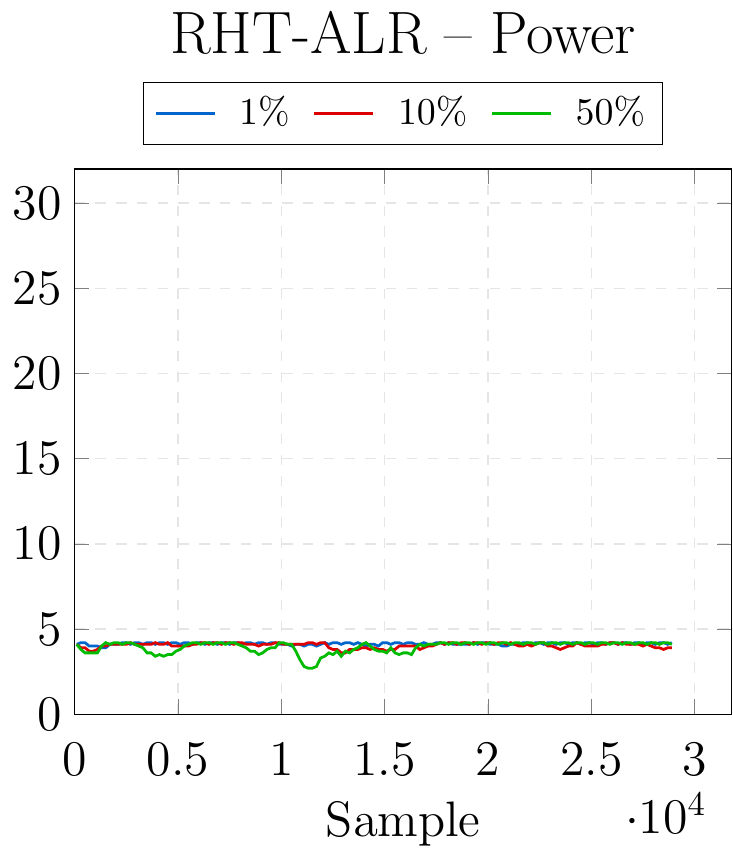}}
	\hspace{0.15cm}
	\subfloat{\includegraphics[width=0.35\columnwidth]{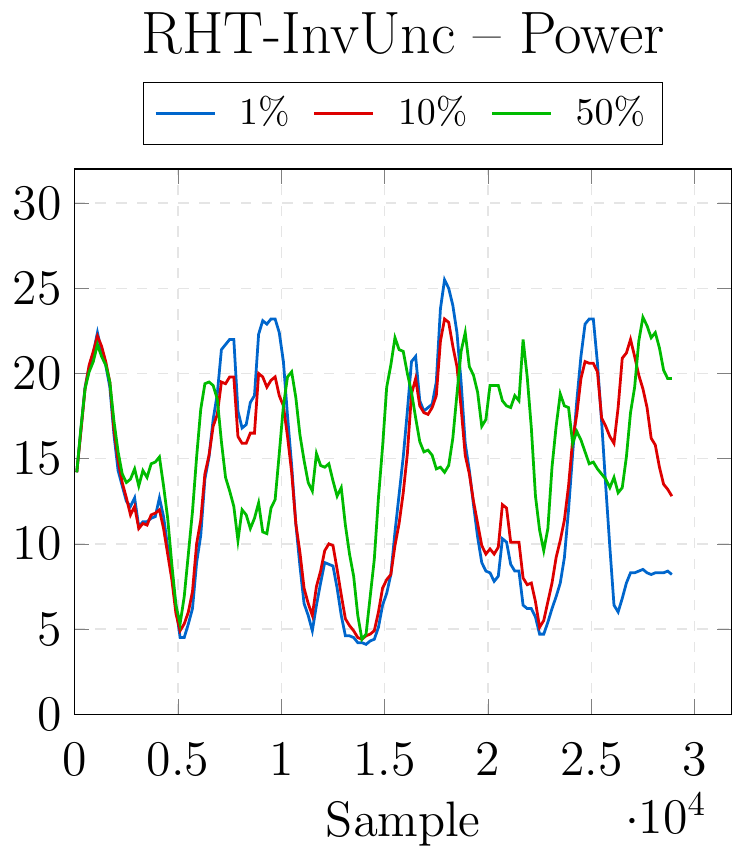}}\\
	\subfloat{\includegraphics[width=0.35\columnwidth]{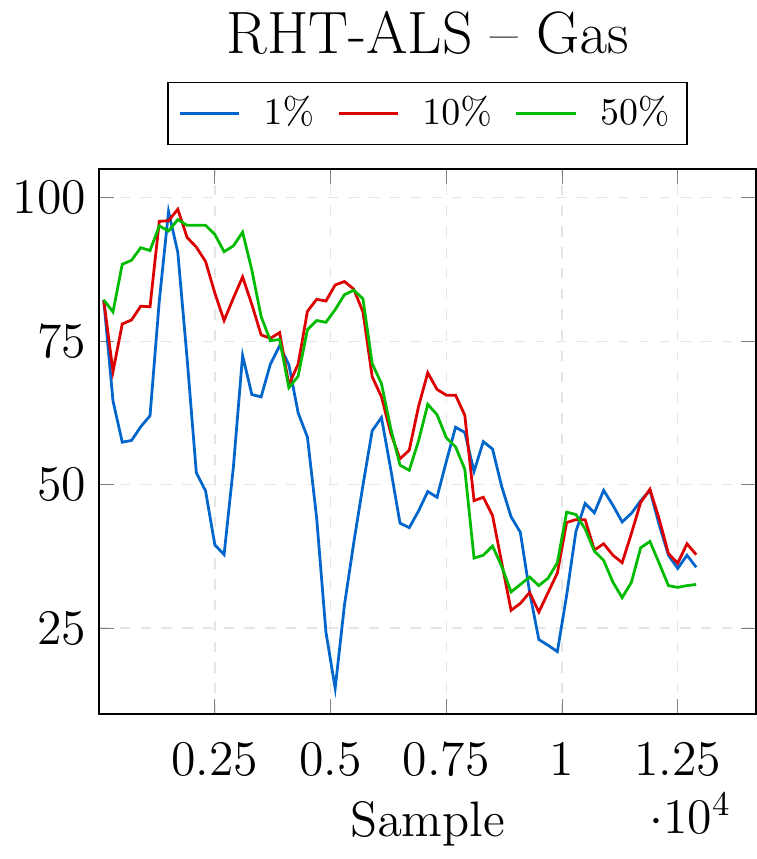}}
	\hspace{0.15cm}
	\subfloat{\includegraphics[width=0.35\columnwidth]{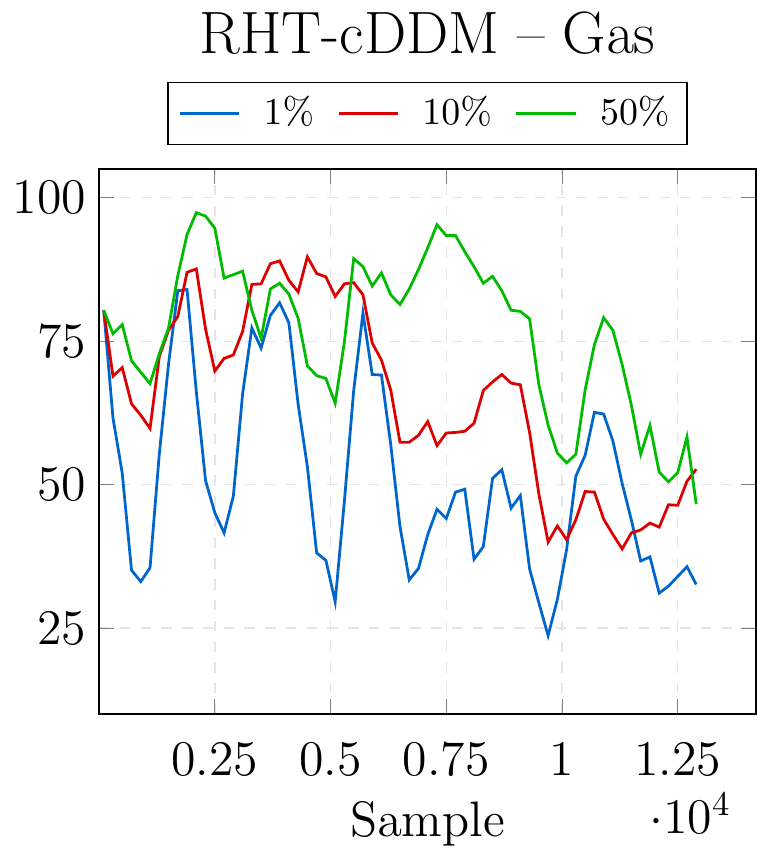}}
	\caption{Accuracy of RHT with different strategies given a budget for the Power and Gas data stream.}
	\label{fig:rht-real}
\end{figure}

In the case of the Power data stream, the hybrid framework using AHT was stable for practically all strategies and budgets, providing some improvements at the same time, mainly for the \texttt{Fixed} and \texttt{WinErr} strategy. However, the boost of performance was rather minor, between 1\% and 2\%. When RHT was used instead of AHT it can be seen that the active learning strategies alone were not capable of maintaining relatively efficient models in the dynamic environment. The accuracy dropped drastically to less than 5\% and in Fig. \ref{fig:rht-real} we can observe that the model is completely useless without a self-labeling module. After adding the \texttt{InvUnc}, \texttt{cDDM}, \texttt{cEDDM} or \texttt{WinErr} strategy the average accuracy rose by about 10\%. In the given example for \texttt{InvUnc} we can see that the algorithm starts reacting to the changes even for very small $B=1\%$.

A few improvements can be noticed for the Gas stream, especially when the informed strategies are combined with both single classifiers. More significant gain is present for RHT and it is another case in which a model using the last four strategies provided significant boosts of accuracy -- from 2\% for \texttt{InvUnc} on $B=10\%$ to almost 15\% for \texttt{cEDDM} on $B=50\%$. In Fig. \ref{fig:rht-real} we can see that the \texttt{cDDM} strategy elevates the accuracy for 10\% and 50\% budget, especially in the second half of the stream. For $B=1\%$ the learning process is slightly more intensified, so it looks a bit more dynamic, however, it does not result in better performance on average. The RCD ensemble improved learning only for the lowest budget. In the case of AWE we can see much more improvements, however, the classifier did not learn the concepts as well as all the rest of the models, being about three times worse than them.

For the Usenet stream, all algorithms worked well on average. Self-labeling strategies with AHT and RCD improved the accuracy only for low budgets up to 10\%, with RHT they did it for the higher ones, and the \texttt{Uni}, \texttt{RandUni} or \texttt{cEDDM} strategy worked well with AWE for all budgets. In the case of the Poker data stream, only AHT and RCD algorithms with the \texttt{Fixed} strategy were able to convincingly enhance the performance for almost all budgets. The ensemble provided a notable gain with all self-labeling strategies on $B=50\%$. For the Elec stream, we can see improvements mainly for committees, however, once again, only RCD performed stably and it can be compared with AHT, which did not cooperate well with self-labeling algorithms, but it achieved the best accuracy on average. On the graph below (Fig. \ref{fig:rcd-real}) we can observe how the \texttt{Fixed} self-labeling strategy improved learning for all presented budgets, for example when at the very beginning the accuracy was elevated above 75\% for $B=10\%$.

Pure active learning approaches on low budgets are sometimes not able to exploit new concepts sufficiently, since there are not enough labeled instances to reinforce a new concept discovery. The hybrid approach that uses a self-labeling step helps with exploiting the concept and creating a more adequate model, without affecting an available budget. This is especially valuable when taking into account that the aim of this approach was to develop methods for learning with very limited access to true class labels. For higher values of the budget, there are a bit fewer improvements, but there are still many of them. Well-modeled concepts may be more easily exploited by self-labeling, since there is a lower risk that an error will be amplified when class boundaries are more or less correct. One must although remember that very large budgets are unrealistic and prohibitive in most real-life scenarios, so we cannot increase it as much as it may be necessary. This is why we did not consider budgets higher than 50\%. On the other hand, when a lot of labeled instances are available, the active learning strategy may be more likely to dominate the learning process and it may be sufficiently good in sampling incoming concepts, while a self-labeling strategy may only impede the process. Such scenarios lead to the situation where the feasibility of a hybrid approach on higher budgets is highly dependent on a stream to which it may be applied.

\subsection{Aggregated results and conclusions}
\label{sec:sum}

Let us summarize the findings from this manuscript and formulate a set of observations and recommendations regarding the usefulness and applicability areas of proposed hybrid methods for mining drifting data streams on a budget. Two following measures are used for this purpose. The first one is the average accuracy (\texttt{Acc}) for all self-labeling strategies over all examined data streams. We calculate it for each classifier and budget separately. The second measure is the fraction (\texttt{Fh}) of cases (each cell for a self-labeling strategy in the result tables is a single case) in which a hybrid approach achieved a better result than any reference method relying only on active learning. The fraction is calculated for each budget and classifier individually. As long as our baseline classifier achieved better than random accuracy, we can assume that improvements are relevant. Therefore, we excluded AWE results for the Sensor and Power stream. The fraction measure tells us if a strategy works well with a given base classifier. The comprehensive summary of all results regarding classifiers, strategies and budgets is presented below.

\begin{figure*}[]
	\centering
	\subfloat{\includegraphics[width=0.5\linewidth]{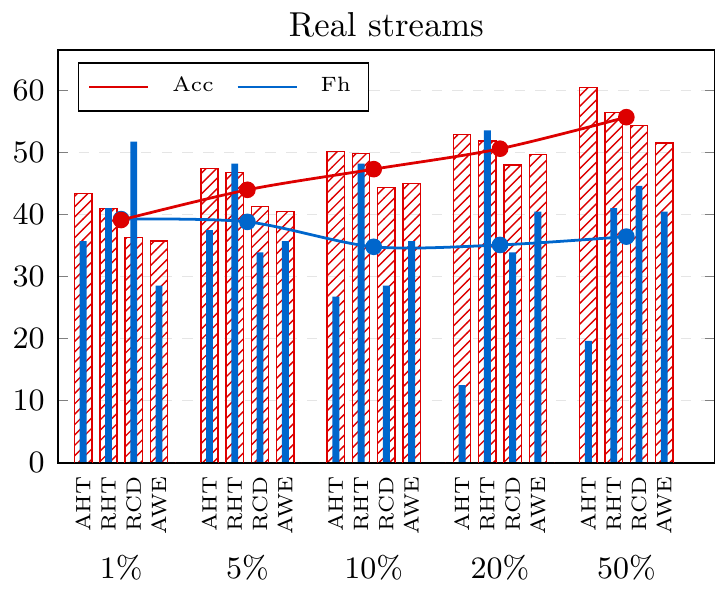}}\hspace*{0.1cm}
	\caption{\texttt{Acc} and \texttt{Fh} for the examined data streams.}
	\label{fig:accfb-cls}
\end{figure*}

\medskip\noindent \textbf{Budget matters}. For real data streams, we can observe that improvements occur for all budgets, but they are present mainly on lower budgets (Fig. \ref{fig:accfb-cls}). The most frequent enhancements can be seen for RCD (in more than a half of cases on $B=1\%$) and RHT (from about 40\% on $B=1\%$ to almost 55\% on $B=20\%$). It is very encouraging since low budgets are the most realistic and practical ones. The RHT classifier reinforced with self-labeling was able to achieve results comparable with very solid AHT. We distinguish the RHT classifier on low budgets as the best for the proposed framework. The good influence of the hybrid approach was also observed for the highest considered budget, on which improvements are relatively frequent. The explanation of why self-labeling works for very low and very high budgets has been presented at the end of Sec.~\ref{sec:realres}. On the other hand, we can notice a local minimum on $B=10\%$ and $B=20\%$. It might mean that the hybrid framework is not good for in-between cases, when models are not very reliable and, at the same time, there are fewer chances for improvements. We can also observe that the average accuracy increases with the budget. This is very intuitive, as more labeled instances support more accurate learning.

\begin{figure*}[]
	\centering
	\subfloat{\includegraphics[width=0.75\linewidth]{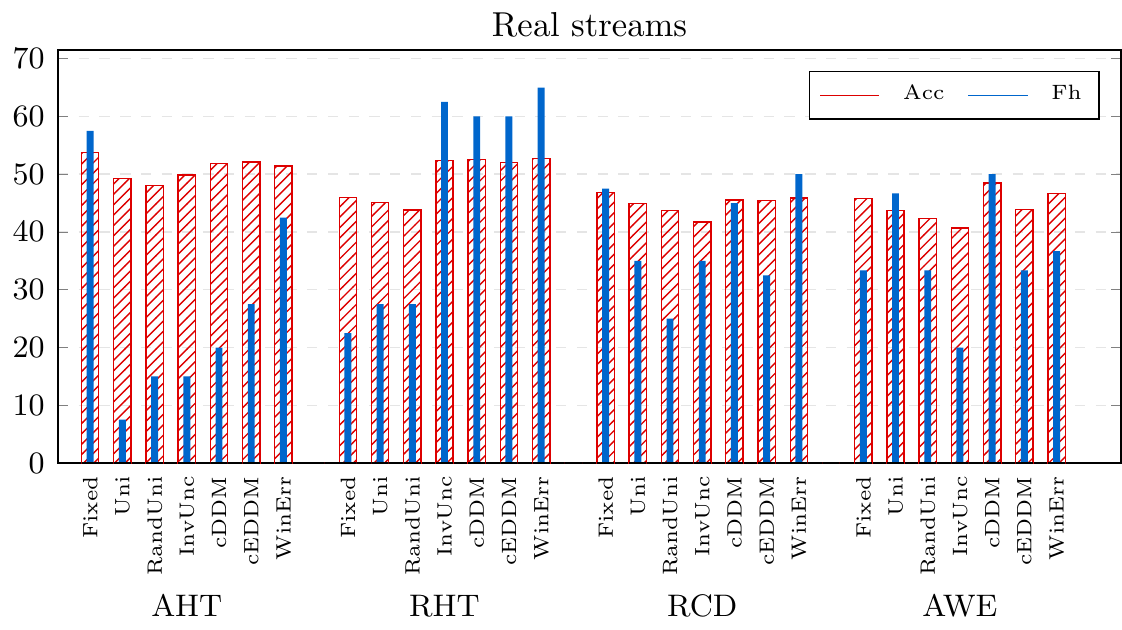}}
	\caption{\texttt{Acc} and \texttt{Fh} for different strategies over the examined data streams.}
	\label{fig:accfb-strategies-real}
\end{figure*}

\medskip\noindent \textbf{Generic wrapper}. We can see that the hybrid framework was able to improve results for many cases and for all of the examined classifiers. For most of the considered real scenarios at least one better hybrid solution could be found (see Sec. \ref{sec:realres}). Since the framework has been designed as a wrapper (a classifier is modular) it is very important that we can observe this fact. One can easily apply the solution to any online classifier in order to boost its performance, especially when facing very limited access to class labels. We recommend experimenting with the framework if results are not satisfying and a classifier cannot be changed. Although the framework is able to improve all considered classifiers, some differences can be noticed. As already mentioned, the highest impact was registered for the RHT classifier. Moreover, both Hoeffding Trees achieved the highest average accuracy in almost all cases (see Fig. \ref{fig:accfb-cls}). In general, single classifiers were able to integrate with self-labeling better than ensembles, which, besides the worse results, are also slower. However, we suppose that committees can be improved in the context of semi-supervised learning. The main reason why they perform worse than single classifiers is probably a fact that for low budgets there are not enough labeled samples to generate sufficiently diversified ensembles. One can notice that the accuracy of the RCD classifier on higher budgets is closer to RHT than on lower ones (Fig. \ref{fig:accfb-cls}).

\medskip\noindent \textbf{Informed over blind}. The informed self-labeling strategies -- \texttt{cDDM}, \texttt{cEDDM} and \texttt{WinErr} -- performed generally better for real streams than the blind approaches, regarding both accuracy and the fraction of enhancements (see Fig. \ref{fig:accfb-strategies-real}). They worked well with all classifiers, but the most significant difference can be observed for the RHT classifier, for which improvements were registered in between 60\% and 65\% of cases. The most straightforward explanation is that the information from drift detectors accordingly and efficiently supported the adjustment of a self-labeling threshold. The only exception is the blind \texttt{Fixed} strategy that uses a very high confidence threshold. It cooperated effectively with AHT and RCD on our examined real-world datasets. The \texttt{InvUnc} strategy was able to significantly improve results only for RHT on real data streams. 

\medskip\noindent \textbf{More for free}. Last but not least, we want to emphasize the most important fact that our hybrid framework can significantly reduce the cost of maintaining online classifiers that work with drifting streaming data. As we could notice, for the Spam data stream the AHT classifier using the active learning strategy without self-labeling was able to achieve nearly 90\% accuracy only if 50\% of objects were labeled. Applying the \texttt{InvUnc} strategy to this case provided a very similar performance with only 1\% of data being annotated, while for the same budget all the active learning strategies were correctly classifying less than 30\% of samples. Let us consider an illustrative example. It has been estimated that more than 500 million tweets is created every day \cite{Twitter:2013stats} and according to the CrowdFlower's offer \cite{Crowdflower:2017site}, annotation of 100 000 rows costs 1500 dollars every month. If one wanted to have 50\% of tweets labeled he would have to spend 6.75 million dollars monthly for that, while for 1\% it is 50 times less, so only 75 000 dollars monthly. This is only a theoretical reduction that self-labeling connected with active learning may provide, but it shows very clearly why the hybrid approach should be considered each time a better accuracy is required and increasing a budget is not feasible.

\section{Summary and future works}
\label{sec:end}

In this paper, we have introduced a novel hybrid framework for learning from drifting data streams on a budget. In real-life scenarios, unlimited access to ground truth cannot be assumed, as the cost is connected with obtaining such information from a domain expert. Therefore, we discussed a set-up of a learning system under scarce access to labels. We have proposed a combination of information coming from active learning and self-labeling, in order to obtain more efficient usage of very few available instances. Active learning allowed for selecting proper ones for label queries, thus leading to the exploration of new concepts emerging from a data stream. These seeds were then utilized by a self-labeling module that offered exploitation of previously discovered structures at no additional costs. We developed two families of algorithms that relied only on classifier outputs or empowered them with additional information from a drift detection module. Seven algorithms were proposed in total, offering a selection between complexity and performance. We recommend applying the hybrid approach if results for all available classifiers are insufficient and one cannot increase a budget or when a weak classifier cannot be replaced.

Experimental analysis showed the usefulness of the proposed hybrid-based approach, especially for realistic scenarios with a highly limited budget. In such cases inclusion of self-labeling allowed to improve the classifier performance by increasing its competence over a discovered concept, while saving the small budget at hand for adapting to changes. We proved that the hybrid framework is a flexible wrapper, so it can work with different classifiers, including online ensembles. We also observed that informed strategies are preferable over blind approaches with an exception to the strategy that uses a high fixed threshold for self-labeling. The proposed hybrid solutions displayed excellent performance for real-life data streams.

The obtained results encourage us to continue our works on learning on a budget from drifting data streams. As a next step, we envision works on another semi-supervised approach which is oversampling. It may be used to enhance the adaptation process by providing more labeled instances without additional cost. We suppose that such an approach may reinforce the construction of diversified online ensembles.

% BibTeX users please use one of
%\bibliographystyle{spbasic}      % basic style, author-year citations
\bibliographystyle{spmpsci}      % mathematics and physical sciences
\bibliography{refs}

\end{document}